\documentclass[sigconf,10pt,nonacm]{acmart}

\title{\system: Flying Drones with Large Language Model}

\author{Guojun Chen}
\email{guojun.chen@yale.edu}
\affiliation{%
  \institution{Yale University}
  \country{}
}

\author{Xiaojing Yu}
\email{xiaojing.yu@yale.edu}
\affiliation{%
  \institution{Yale University}
  \country{}
}

\author{Neiwen Ling}
\email{neiwen.ling@yale.edu}
\affiliation{%
  \institution{Yale University}
  \country{}
}

\author{Lin Zhong}
\email{lin.zhong@yale.edu}
\affiliation{%
  \institution{Yale University}
  \country{}
}
\usepackage{array}
\usepackage{colortbl}
\usepackage{xurl}
\usepackage{color}
\usepackage{soul}
\usepackage{siunitx}
\usepackage{listings}
\usepackage{subcaption}
\usepackage{float}
\usepackage[normalem]{ulem}
\usepackage{setspace}
\usepackage{placeins}
\usepackage{afterpage}
\usepackage{hhline}
\usepackage{enumitem}
\usepackage{pifont}
\usepackage{arydshln}
\usepackage{adjustbox}
\usepackage{lipsum}
\usepackage{tabularx}
\usepackage{makecell}
\usepackage{wrapfig}
\usepackage{moreverb}
\usepackage{multirow}
\usepackage{xspace}
\usepackage{hyperref}
\usepackage{epsfig}
\usepackage{endnotes}
\usepackage{url}
\usepackage{fancyhdr}
\usepackage{cascadia-code}
\usepackage{xcolor}
\usepackage{caption}
\usepackage[framemethod=TikZ]{mdframed}
\definecolor{codegray}{RGB}{128, 128, 128}
\definecolor{backcolor}{RGB}{250, 250, 250}
\definecolor{rulecolor}{RGB}{120, 120, 120}
\definecolor{keywordcolor}{RGB}{218, 75, 88}
\definecolor{stringcolor}{RGB}{64, 135, 39}
\definecolor{identifiercolor}{RGB}{50, 50, 50}

\usepackage{DejaVuSansMono}
\lstdefinestyle{mystyle} {
    commentstyle=\color{rulecolor},
    keywordstyle=\color{keywordcolor},
    stringstyle=\color{stringcolor},
    basicstyle=\ttfamily\scriptsize,
    identifierstyle=\color{identifiercolor},
    backgroundcolor=\color{backcolor},
    breakatwhitespace=false,
    breakindent=0pt,
    breaklines=true,
    captionpos=b,
    keepspaces=true,
    numbers=left,
    numberstyle=\ttfamily\tiny\color{rulecolor},
    numbersep=3pt,
    showspaces=false,
    showstringspaces=false,
    showtabs=false,
    framexleftmargin=0pt,
    frame=lines,
    rulecolor=\color{rulecolor},
    rulesepcolor=\color{gray},
    xleftmargin=0pt,
    xrightmargin=0pt,
}
\definecolor{blueframecolor}{RGB}{173,212,251} 
\definecolor{textgray}{gray}{0.5}
\definecolor{typeflyblue}{RGB}{95,148,247} 
\definecolor{typeflyred}{RGB}{202,85,92}
\definecolor{typeflygreen}{RGB}{82,133,54}
\definecolor{typeflypurple}{RGB}{189,21,240}
\newmdenv[  
  linecolor=blueframecolor,
  outerlinewidth=0.5pt, % 边框线加粗
  roundcorner=3pt, % 增加圆角的半径来使其更明显
  backgroundcolor=blueframecolor!10,
  frametitlerule=true,
  innertopmargin=3pt, % 调整这里以减少顶部内边距
  innerbottommargin=3pt,
  font=\small
]{customframe}

\newcommand{\lineref}[1]{line~\autoref{#1}}

\newcommand{\lstref}[1]{Listing~\ref{#1}}

\definecolor{lightgray}{gray}{0.9}
\definecolor{lightblue}{rgb}{0.9,0.9,1}
\definecolor{blue_bg}{rgb}{0.85,0.85,1}
\definecolor{lightyellow}{rgb}{1,1,0.8}
\definecolor{lightpurple}{rgb}{1,0.85,1}
\definecolor{red}{rgb}{1,0,0}
\definecolor{darkgreen}{rgb}{0.4,0.7,0.3}

\newcommand*\wcircled[1]{
  \tikz[baseline = (char.base)]{
    \node[shape = circle, draw = black, inner sep = 0.5pt] (char) {\textcolor{black}{\textit{#1}}};
  }
}

\usepackage{tcolorbox}
\newcommand{\program}[1]{\textsf{#1}}
\newcommand{\prog}[1]{\textsf{#1}}
\newcommand{\code}[1]{\texttt{\footnotesize #1}}

\newcommand{\task}[1]{\textit{\textcolor{typeflygreen}{#1}}}

\newcommand{\hlc}[2][yellow]{ {\sethlcolor{#1} \hl{#2}} }

\newcommand\lin[1]{\hlc[yellow]{LZ: -- #1 --}}

\newcommand{\remove}[1]{}

\newcommand{\system}[0]{\textsf{TypeFly}\xspace}
\newcommand{\llmc}[0]{\textsf{Prompt Generator}\xspace}
\newcommand{\llm}[0]{\textsf{Remote LLM}\xspace}

\newcommand{\lang}[0]{\textsf{MiniSpec}\xspace}
\newcommand{\langI}[0]{\textsf{MiniSpec Runtime}\xspace}
\newcommand{\spskill}[0]{\textsf{probe}\xspace}
\newcommand{\replan}[0]{\textsf{replan}\xspace}

\newcommand{\ve}[0]{Vision Encoder\xspace}
\newcommand{\plan}[0]{plan\xspace}

\newcommand{\plans}[0]{plans\xspace}
\newcommand{\skill}[0]{skill\xspace}

\newcommand{\skills}[0]{skills\xspace}
\newcommand{\Skills}[0]{Skills\xspace}

\newcommand{\taskd}[0]{task description\xspace}

\newcommand{\scened}[0]{scene description\xspace}
\newcommand{\sceneds}[0]{scene descriptions\xspace}

\newcommand{\exemode}[0]{\textit{Stream Interpreting}\xspace}

\begin{document}
\begin{abstract}
% \nl{since the paper focus on the design of MiniSpec, shall we emphasize it in the title, like "A latency-efficeint language for LLM-controlled drone"? }
% \lin{It depends on the final story. If it is about ChatFly as a system with MiniSpec being a component, the current title might be good. Maybe wait after we finalize the intro and abstract to revisit the title.}
% Commanding a robot with a natural language is not only user-friendly but also opens the door for emerging language agents to control the robot. 
% Emerging large-language models (LLMs) provide a previously impossible opportunity to automatically translate a task description in a natural language to a program that can be executed by the drone. However, the most powerful LLMs and their vision counterparts are limited in three important ways. 
% First, they are slow, the latency costs proportionally to the request and response lengths.
% Second, they are only available as cloud-based services. Sending images to the cloud raises privacy concerns. Finally, without expensive fine-tuning, existing LLMs are quite limited in their ability to write a program for specialized systems like drones in a succinct manner.

Recent advancements in robot control using large language models (LLMs) have demonstrated significant potential, primarily due to LLMs' capabilities to understand natural language commands and generate executable plans in various languages.
However, in real-time and interactive applications involving mobile robots, particularly drones, the sequential token generation process inherent to LLMs introduces substantial latency, i.e. response time, in control plan generation. 
%This latency adversely impacts both system responsiveness and user experience.
%This challenge highlights the necessity for optimization strategies to reduce the wait time for plan generation, as current approaches primarily focus on plan accuracy without adequately improving response time.

In this paper, we present a system called \system that tackles this problem using a combination of a novel programming language called \lang and its runtime to reduce the plan generation time and drone response time. That is, instead of asking an LLM to write a program (robotic plan) in the popular but verbose Python, \system gets it to do it in \lang specially designed for token efficiency and stream interpretation. Using a set of challenging drone tasks, we show that design choices made by \system can reduce up to $62\%$ response time and provide a more consistent user experience, enabling responsive and intelligent LLM-based drone control with efficient completion.
\end{abstract}

\maketitle

\section{INTRODUCTION}
\label{sec:intro}

Mobile robots, especially drones, have seen growing applications in both personal and public domains. Recently, the robotics community has leveraged the remarkable capabilities of large language models (LLMs) and their multi-modal counterparts (Vision Language Models) for understanding task descriptions in natural language and generating plans in the form of predefined function calls or Python programs~\cite{driess2023arxiv,mirchandani2023arxiv,vemprala2023arxiv,huang2023arxiv,wu2023arxiv,tazir2023wcse}, beating traditional robotic control methods.

In these applications, the user experience heavily depends on how quickly the robot can start to act after receiving the user's commands and complete the task~\cite{pelikan2023delay}. Unfortunately, the sequential token generation of LLMs leads to inference latency that is proportional to the length of the output plan, increasing both the response and completion times of the robot control~\cite{liu2023deja}. 
% This poses challenges for real-time applications of LLMs in robot control. Current work primarily focuses on the correctness of plans, but effective optimization methods for real-time requirements are still lacking.

% \gca{bring response time in the first paragraph}
% \nl{since we focus on reducing the latency now, we may also need to discuss what's the future meaning of this paper when their emerges more light-weight LLM. One solution is clarifying from the aspect of stream interpreting}

\begin{figure*}[t]
  \centering
  \includegraphics[width=1\textwidth]{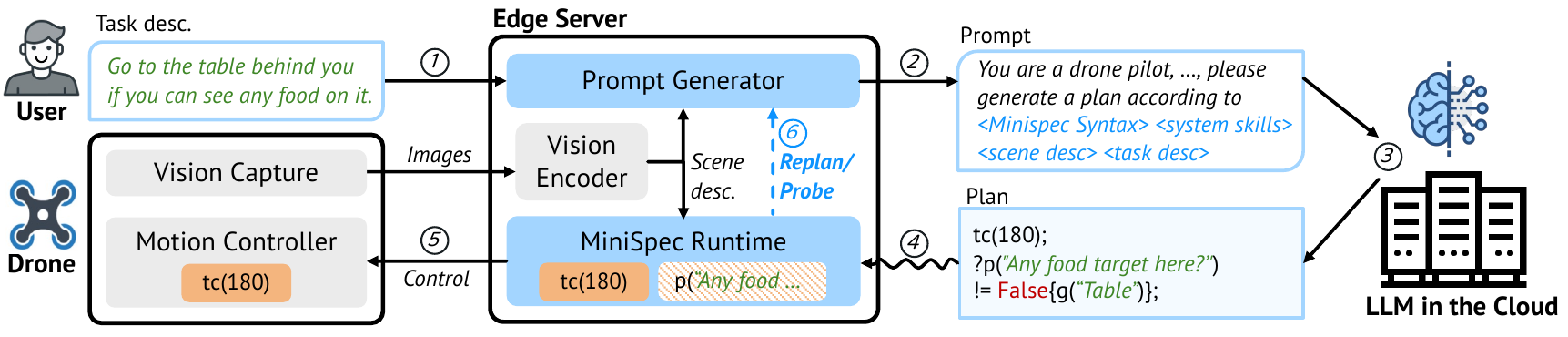}
  % \vspace{-2ex}
  \caption{\textmd{System overview of \system: An on-premise edge server controls the drone to accomplish a task described by a user, whether a human or a language agent, using natural language. Based on the task description and a scene description by a vision encoder, the LLM writes a program in \lang (\S\ref{sec:language_design}), called \plan. In this example, the \plan includes one elementary statement \code{tc(180)}, which turns the drone by 180-degree, and one composite statement \code{?p("Any food target here?")!=False\{g("Table")\}}, which moves the drone to a table, depending on whether there is food on it. With \exemode (\S\ref{sec:runtime}), the drone can start to act on a statement while \llm is still generating the next. \system can deal with syntax errors and unexpected situations through \lang's exception handling mechanism \code{replan}(\S\ref{sec:replan}). Additionally, using a special skill \spskill (\S\ref{sec:system-skills}), \system can engage the LLM during the plan execution.}}
 % \lina{This caption needs some thoughts. It helps the readers to understand how the system works but does not explicitly cover all the key points: (1) LLM programs in MiniSpec; (2) exception handling by replan; (3) probe as a special skill; (4) stream interpretation. The caption covers (4) but not explicitly (1) to (3)... }
  \label{fig:system-diagrams}
  % \lina{Where are replan and probe? (6) upward?}
  % \lin{The output of LLM (plan) should go directly to MiniSpec Interpreter, not through LLMController. Likewise (6) should be unidirectional, upward, not bidirectional.}
  % \gc{any comment on the interpreting mode?}
  % \lin{(5) arrow should go to LLMController. No need to visually show these modes.}
  % \lin{MiniSpec Interpreter should be MiniSpec Runtime, which can compile or interpret the plan....}
\end{figure*}

This paper presents \system, an end-to-end system that enables an LLM  to efficiently accomplish complicated tasks with low latency, as described by \autoref{fig:system-diagrams}. 
The key idea of \system is to design a small, special programming language, called \lang, for the LLM to write drone control plans with high token efficiency and chance of success, as compared to the popular choice of Python.
As illustrated by \autoref{fig:system-diagrams}), a \system user provides a \taskd in English (\wcircled{1}); then the \llmc service combines the \taskd and \scened (generated by \ve based on video stream from the drone, in English) into a Planning prompt (\wcircled{2}) and send it to the \llm (\wcircled{3}). 
The \llm responds, in streaming generation mode, with a \plan, which is interpreted by the \langI (\wcircled{4}) and carried out by the drone (\wcircled{5}). During the interpretation and execution of the \plan, the \langI may engage the \llm through the \llmc (\wcircled{6}), using a special system skill \spskill and  \lang's exceptional handling mechanism, namely \code{replan}.

Combining innovations with best practices from the literature, \system achieves four crucial goals for drone applications for the first time.
(\textit{i}) \textit{Efficient Planning}. By using \lang and its \exemode, \system minimizes the number of plan tokens and the response time. 
(\textit{ii}) \textit{Efficient Execution}. \system reduces the task completion time by integrating LLM during execution, using a special skill called \spskill.
(\textit{iii}) \textit{Privacy}. \system leverages an on-premise edge server to process the images into text descriptions and only send such text descriptions to the \llm service.
(\textit{iv}) \textit{Safety}. The design of \lang forbids infinite loops and the use of third-party libraries in the \plan output by the LLM.

We overview \system in \S\ref{sec:design} and describe the design of \lang in \S\ref{sec:minispec}, and our prototype implementation in \S\ref{sec:implementation}.
In \S\ref{sec:eval}, we report an experimental evaluation of \system using the prototype and a benchmark of 11 tasks of various levels of complexity. Our evaluation shows that the use of \lang can reduce the response time up to $62\%$ and provide relatively consistent performance (<$1.5s$ response time for all tasks) across tasks with different complexities, compared to the popular choice Python. It also shows that the mechanism of \spskill and \lang exception handling is critical to its capabilities of accomplishing complex tasks, which are beyond the reach of any existing systems.
Our evaluation also reveals intriguing limits of \system, especially its use of \prog{GPT4}, which suggests future work (\S\ref{sec:concluding}).
%Be defensive about potential criticism that we did not consider usability.
While the user plays the key role of providing a precise and correct \taskd in \system, we focus on the systems aspects of \system in this paper and leave its usability out of the scope. 

In summary, this paper makes the following contributions:
\begin{itemize}[leftmargin=1em]
     \item \system, an end-to-end latency-efficient system that flies a drone to accomplish complex tasks described in English.
     
     \item \lang, a small programming language designed for LLMs to generate robotic task \plans efficiently. \lang features a special skill \code{probe} to simplify the logic and improve the effectiveness of the plan. \lang also has a special keyword \code{replan} to handle exceptions such as syntax errors and unexpected situations.
     \item \exemode, streaming generation and in-time interpreting execution for LLM-powered drone control, which significantly reduces the response time of the drone. Response time refers to the duration from the user query to the first action of the drone. To the best of our knowledge, we are also the first to use this generation-execution mode in robotic control.

\end{itemize}
The source code of \system along with prompts and video recordings of evaluation will be made openly available.

\section{Background}
\label{sec:background}
% \lina{Add a small paragraph of introductory material for LLM: how it is trained, its architecture, how it has been growing in size and capability in the past few years, how it has changed many fields.....Assume the reader only heard of LLM in the news. This is your chance to help them learn a little more.}

% Trained on massive textual data including books, codebases, and websites, LLMs have demonstrated unprecedented capabilities in language understanding and generation, representing a significant advancement in AI. 
% Such capabilities have seen a dramatic increase as LLMs grow in size, from 1.5 million parameters in \prog{GPT2}, to over 175 billion in \prog{GPT3}, and over a trillion in \prog{GPT4} according to unofficial speculations~\cite{George2023youtube}, over merely 4 years.

The capabilities of LLMs have dramatically increased with their growing size.
\prog{GPT4}, one of the largest models and best general-purpose LLMs, excels particularly in reasoning for math, algorithms, and code generation~\cite{achiam2023gpt}.
Serving large models like \prog{GPT4} requires massive computational resources, typically involving numerous network-connected GPU servers in data centers, so the most capable LLMs are usually available as cloud-based services.
Despite intensive research on training or fine-tuning smaller LLMs for edge-based deployment, these models still lag far behind the capabilities of cloud-based LLMs.
For example, according to~\cite{zhou2023language}, \prog{GPT4} shows an unprecedented 94\% Pass@1 rate on the HumanEval~\cite{chen2021evaluating} dataset, as a comparison, the best local deployable model Code Llama~\cite{rozière2023code} only achieves a 62\% Pass@1 rate. 
Therefore, in this work, we opt for cloud-based \prog{GPT4} in designing and implementing \system, to tap the latest capabilities of LLMs. Nevertheless, much of \system's design is agnostic of where the LLM is served, cloud or edge. 

\begin{figure}[t]
    \centering
    \includegraphics[width=0.480\textwidth]{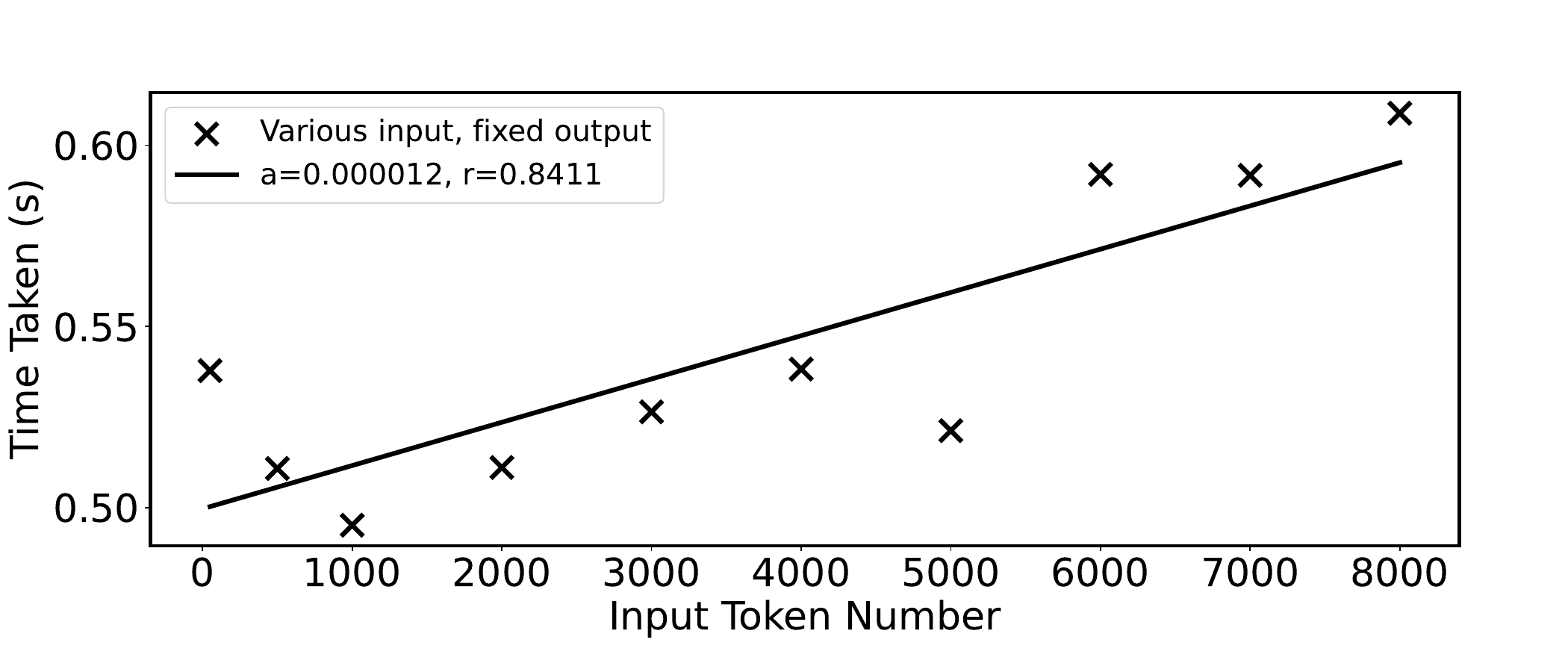}
    \includegraphics[width=0.480\textwidth]{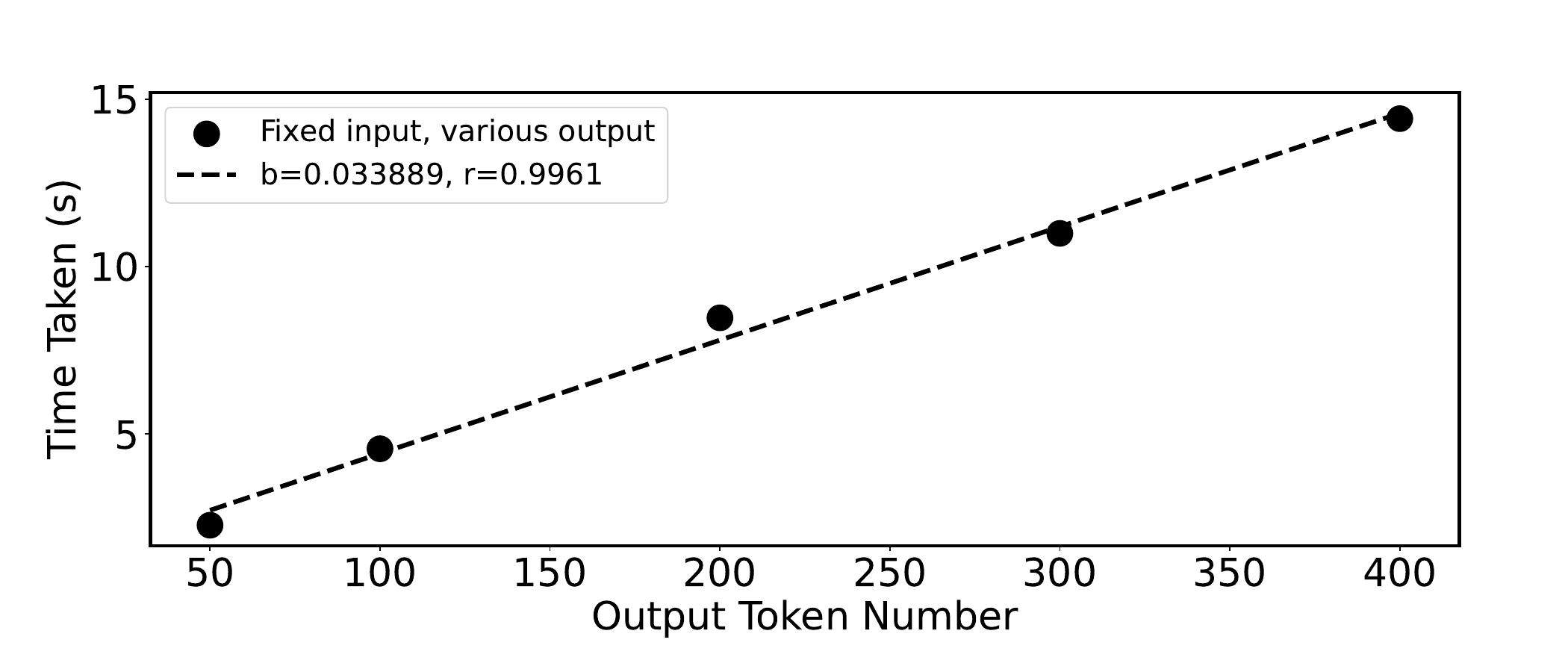}
    % \vspace{-2ex}
    \caption{\textmd{GPT4 API latency regards different input and output token numbers (each point is the average of 10 measurements). 
    The top figure represents measurements taken with changing input token numbers while keeping the output token constant; The bottom figure is measured with various output token numbers and fixed input token numbers. The trend lines suggest that $b\approx 2800a$.
    Despite the low correlation coefficient between latency and the number of input tokens, we can still conclude an estimation that generating output tokens is more than 1000 times slower than processing input tokens. (The measurements were conducted on March 5, 2024, using \code{gpt-4} model)}}
    \label{fig:gpt4-latency}
\end{figure}

\paragraph{Latency} The latency of using a remote LLM such as GPT4 consists of that from both the network and computation.
At the time of this writing, the computation latency is larger than the network's (10s of milliseconds) by at least an order of magnitude for only 5 output tokens. Our experience and measurement indicate that the latency of using GPT4 can be closely approximated with the following formula:
\[ latency = a\cdot N_p+b\cdot N_o+c\]
where $N_p$ and $N_o$ are the number of input and output tokens, respectively. They can be measured by the length of the encoded input (prompt) and output text, using the OpenAI tiktoken tokenizer \cite{openai2023tiktoken}. $c$ is the network latency and service schedule delay. As shown in \autoref{fig:gpt4-latency}, $b\approx 2800a$ during our measurement. This implies that the number of output tokens has a major impact on the total latency, provided that the count of input tokens is not significantly (more than 2800 times) larger than that of the output tokens. Based on this finding, \system employs a long-prompt short-output strategy to minimize the required token number for planning (detailed in \autoref{sec:implementation}).
%\nl{we used to mention it in the section of prompt generation, but now this part has been deleted}

\paragraph{Response Mode}
LLMs usually support two modes for generating responses: batch generation and streaming generation. In batch generation, the LLM returns the entire response at once. This can be useful for short, concise answers or when the complete context is needed for further processing. However, it can be less responsive for longer outputs as the user must wait until the entire response is generated. In streaming generation, the LLM generates the response incrementally, sending parts of the response to the user as they are generated. This creates the effect of a continuous, flowing output and improves responsiveness as well as user engagement. \system leverages the streaming generation and the design of \lang to reduce the planning response time.

\section{Related Work}
\label{sec:related}
%As discussed in \S\ref{sec:background}, LLM-based plan generation introduces linear latency costs with the plan token count. 

Related to \system's end goal of commanding a drone with a natural language, there is a rich literature from the robotics community that apply NLP to robotic control, e.g., ~\cite{blukis2019learning,craparo2004natural}.
% \system is just one of the many systems that have emerged in the past year or so that apply LLMs in robotic control, achieving much better natural language understanding and much more flexible planning~\cite{vemprala2023arxiv,huang2023arxiv,wu2023arxiv,liang2023icra,singh2023icra,driess2023arxiv,zitkovich2023crl,tazir2023wcse,huang2023grounded,shah2023lm}. 
\system is one of many systems that have emerged in the past year applying LLMs to robotic control, achieving significantly better natural language understanding and more flexible planning from task descriptions in natural language ~\cite{vemprala2023arxiv,huang2023arxiv,wu2023arxiv,liang2023icra,singh2023icra,driess2023arxiv,zitkovich2023crl,tazir2023wcse,huang2023grounded,shah2023lm} or formal robotic languages like PDDL~\cite{silver2022pddl,silver2024generalized}.
% Due to space limit, we will focus on related systems that also leverage LLMs to generate plans. 
% Compared to these systems, \system stands out in accomplishing complex tasks and its set of innovations to reduce the latency.
Due to space limitations, we will focus on related systems that leverage LLMs to generate program-based plans that stand out by enabling more complex tasks with advanced condition evaluations and loops beyond sequential logic.
\system introduces latency-reducing innovations while retaining the flexibility of program generation methods.

A key innovation of \system is to design a programming language, i.e., \lang, for LLMs to write programs, which reduces the latency and program errors. 
To the best of our knowledge, \system is the first to leverage a custom language to optimize the response time when facilitating LLMs in robotic planning.
In contrast,  most related systems use Python~\cite{vemprala2023arxiv,huang2023arxiv,wu2023arxiv,liang2023icra,singh2023icra}, due to its extensive corpus on which most public LLMs were trained.
Although Python is designed to be concise, it is intended for humans to write programs. As a result, Python programs are not token-efficient.
In contrast, \lang is designed for LLMs and would use on average 30\% fewer tokens for the same semantics.
Moreover, using Python could introduce unexpected use of third-party libraries or infinite loops, which may not be safe. 
In contrast, \lang forbids infinite loops and no existing LLMs were trained with \lang libraries (yet). 
% One work uses Keyhole Markup Language (KML)~\cite{qifei2024edge} language for flight control in a mission planner like QGroundControl~\cite{qgc}, though focus only on planning capabilities. 
One work uses Keyhole Markup Language (KML)~\cite{qifei2024edge} language for flight control in a mission planner like QGroundControl~\cite{qgc}. However, this approach only focuses on planning capabilities. And execution must take place after the entire plan has been generated, as it needs to be fully uploaded to the QGroundControl software for implementation.

\system leverages \lang's exception handling mechanism, \code{replan}, to implement incremental planning (\S\ref{sec:replan}), which is related to but more efficient than full online planning that is widely used in robotics~\cite{garrett2020online, nageli2017realtime}. Online planning generates a plan for each step of the plan based on the current scene. In contrast, in \system, \code{replan} is invoked only when it is necessary.

Another key innovation of \system is to engage the LLM in the execution of a \plan via the special skill \spskill (\S\ref{sec:system-skills}). Related, the authors of~\cite{song2023arxiv} also involve the LLM throughout the whole execution phase by asking the LLM to generate a new plan after every several actions, which is both slow and expensive.
In contrast, \system incorporates the LLM in a novel way: it steps in to provide responses when confronted with human language requests that the vision model cannot comprehend. This significantly enhances the reasoning capabilities and improves the efficiency of task execution to adapt to a dynamic environment.

Related to \system's support of stream interpretation of \lang plans, the LLM Compiler~\cite{kim2023llm} asynchronously executes functions and generates programs by analyzing dependencies between variables. However, Python’s complex syntax tree makes syntax fragmentation extremely challenging, limiting the LLM Compiler to simple sequential execution scenarios.
In contrast, \lang allows \system to asynchronously plan generation and execution (even in loop conditions),  significantly reducing both response and task completion latencies.

\section{System Overview}
\label{sec:design}

\system is a low-latency, end-to-end system that leverages LLM to control the drone via a custom program language \lang, enabling it to accomplish complex tasks with quick response efficiently. In designing \system, we aimed to leverage both the edge computing resources and cloud-based LLM capabilities, to successfully handle non-predefined natural language tasks commanded by users while protecting user's privacy.

As shown in \autoref{fig:system-diagrams}, 
a drone with the motion controller and video capture capability, \lang-related modules running in an on-premise edge server, and a cloud-powered LLM. The drone sends camera capture to the edge server via a WiFi connection. The user, either a human or a language agent, can observe real-time video feedback and provide high-level instructions to the edge server in a natural language to \system(\wcircled{1}), called \taskd. Simultaneously, the vision encoder on the local server provides a \scened API for other modules based on these images.

The edge server plays a central role in generating effective and efficient \lang by calling the LLM and performing interpreting functions for robot control, utilizing two key components.
First, the \llmc sends a prompt to the \llm consisting of the syntax of \lang, \taskd, \scened, and other relevant information and examples about \system to help formulate executable solutions written in \lang that meet the task objectives (\wcircled{2}).
Second, \langI, which is responsible for parsing and executing the \lang plan received from the \llm (\wcircled{3}). When \langI receives the generated token from the \llm in streaming mode (\wcircled{4}), it will execute the \lang plan immediately after a statement is generated and parsed executable. Simultaneously, \langI continuously receives the subsequent parts of the plan from \llm.

During the plan interpretation and drone control (\wcircled{5}), the edge services may also engage the \llm through the \llmc (\wcircled{6}) by executing a special system skill \spskill for efficient query and using the keyword \code{replan} in \lang for exceptional handling.

\textit{System's scalability.} While we implement the \system design for miniature drones, our \system is portable to other robotic platforms: only some of the system skills (Drone Control and High-level skills) are platform-specific, while the rest of \system, including \lang, \llmc and \ve, are platform-agnostic.  

\textit{User's Role.} The user, either human or language agent, interacts with the \system using a natural language. The user provides the \taskd, receives the \code{log} output, and observes the drone's real-time video capture as the \system carries out the task. Additionally, the user can terminate the execution at any time. 
% \sout{Even though the LLM can help (See \S\ref{sec:semantics}), it is the user who is ultimately responsible for confirming the semantic accuracy of the plan and ascertaining if the task has been correctly completed.}\nl{delete this sentence?}

\begin{figure*}[t]
% \begin{multicols}{2}
\begin{lstlisting}[label={minispec-syntax}, caption={\textmd{\small Backus–Naur Form Syntax of \lang. MiniSpec meets the all criteria for Turing completeness except the infinite loop, which is intentionally forbidden at the language level. Only system skills can be called as a function with MiniSpec, preventing potential breaks to system integrity. \code{replan} as the exception handling mechanism can be actively inserted into the plan as a statement, enabling recovery from unexpected situations for improving the success rate.}},escapeinside={(*@}{@*)}]
(*@\color{typeflyred}{<program>}@*) ::= { <composite-statement> [(*@\color{typeflygreen}{`;'}@*)] | <statement> (*@\color{typeflygreen}{`;'}@*) } (*@\hfill@*) (*@\color{typeflyblue}{\% Program structure
}@*)

(*@\color{typeflyred}{<statement>}@*) ::= <function-call> | <variable-assign> | <exception-handling> | <return>(*@\hfill@*) (*@\color{typeflyblue}{\% Different statements
}@*)
(*@\color{typeflyred}{<function-call>}@*) ::= <function-name> [(*@\color{typeflygreen}{`('}@*) <argument> (*@\color{typeflygreen}{`)'}@*)] (*@\hfill@*) (*@\color{typeflyblue}{\% Function call (search an apple):} \color{typeflypurple}{s(`apple')}@*)
(*@\color{typeflyred}{<function-name>}@*) ::= (*@\color{typeflygreen}{`s'}@*) | (*@\color{typeflygreen}{`sa'}@*) | (*@\color{typeflygreen}{`tc'}@*) | (*@\color{typeflygreen}{`p'}@*) | (*@\color{typeflygreen}{`mf'}@*) | ... (*@\hfill@*) (*@\color{typeflyblue}{\% Function names can only be system skills}@*)
(*@\color{typeflyred}{<argument>}@*) ::= <value> { (*@\color{typeflygreen}{`,'}@*) <value> } (*@\hfill@*) (*@\color{typeflyblue}{\% Function arguments}@*)
(*@\color{typeflyred}{<variable-assign>}@*) ::= <variable> (*@\color{typeflygreen}{`='}@*) <function-call> (*@\hfill@*) (*@\color{typeflyblue}{\% Variable assignment (save in variable }\color{typeflypurple}{\_1}\color{typeflyblue}{):} \color{typeflypurple}{\_1=s(`apple')}@*)
(*@\color{typeflyred}{<exception-handling>}@*) ::= (*@\color{typeflygreen}{`replan'}@*) | (*@\color{typeflygreen}{`rp'}@*) (*@\hfill@*) (*@\color{typeflyblue}{\% Exception handling}@*)
(*@\color{typeflyred}{<return>}@*) ::= (*@\color{typeflygreen}{`->'}@*) <value> (*@\hfill@*) (*@\color{typeflyblue}{\% Return statement}@*)

(*@\color{typeflyred}{<composite-statement>}@*) ::= <loop> | <conditional> (*@\hfill@*) (*@\color{typeflyblue}{\% Block statements}@*)
(*@\color{typeflyred}{<loop>}@*) ::= <int> (*@\color{typeflygreen}{`\{'}@*) <program> (*@\color{typeflygreen}{`\}'}@*) (*@\hfill@*) (*@\color{typeflyblue}{\% Loop structure (loop for 10 times):} \color{typeflypurple}{10\{...\}}@*)
(*@\color{typeflyred}{<conditional>}@*) ::= (*@\color{typeflygreen}{`?'}@*) <condition> (*@\color{typeflygreen}{`\{'}@*) <program> (*@\color{typeflygreen}{`\}'}@*) (*@\hfill@*) (*@\color{typeflyblue}{\%  Condition stmt. (if found apple, then do...):} \color{typeflypurple}{?s(`apple')\{...\}}@*)
(*@\color{typeflyred}{<condition>}@*) ::= <operand> <comparator> <operand> { (*@\color{typeflygreen}{`\&'}@*) <condition> | (*@\color{typeflygreen}{`|'}@*) <condition> } (*@\hfill@*) (*@\color{typeflyblue}{\% Condition structure}@*)
(*@\color{typeflyred}{<operand>}@*) ::= <value> | <function-call> (*@\hfill@*) (*@\color{typeflyblue}{\% Operands in conditions}@*)
(*@\color{typeflyred}{<comparator>}@*) ::= (*@\color{typeflygreen}{`>'}@*) | (*@\color{typeflygreen}{`<'}@*) | (*@\color{typeflygreen}{`=='}@*) | (*@\color{typeflygreen}{`!='}@*) (*@\hfill@*) (*@\color{typeflyblue}{\% Comparators}@*)

(*@\color{typeflyred}{<value>}@*) ::= <literal-value> | <variable> (*@\hfill@*) (*@\color{typeflyblue}{\% Values in the language}@*)
(*@\color{typeflyred}{<variable>}@*) ::= (*@\color{typeflygreen}{`\_'}@*) <int> (*@\hfill@*) (*@\color{typeflyblue}{\% Variables}@*)
(*@\color{typeflyred}{<literal-value>}@*) ::= <int> | <float> | <string> | <bool> (*@\hfill@*) (*@\color{typeflyblue}{\% Literal values}@*)
\end{lstlisting}

% \end{multicols}
% \vspace{-1em}
\end{figure*}

\textit{Vision Analysis: LLM vs. VLM.}~~
We opt for a combination of LLM and lightweight vision encoders instead of leveraging the latest advancements in visual-language models (VLMs)~\cite{reed2022arxiv,jiang2023arxiv,driess2023arxiv,zitkovich2023crl}. 
Similar combinations have also been adopted by others~\cite{huang2023arxiv,wu2023arxiv,singh2023icra,lin2023arxiv,zhao2023arxiv}. 
Deploying VLMs directly on edge servers incurs significant computational overhead and latency while deploying them in the cloud raises privacy concerns due to potential reverse-engineering of image embeddings. Our strategy reduces the LLM service cost and enhances privacy by only sending language-based descriptions of the environment to the cloud.

\section{\lang}
\label{sec:minispec}

A key idea embraced by \system is to design its own language for LLMs to write programs (plans). 
We eschew Python, a popular choice for related work, for reasons discussed in \S\ref{sec:related}. 
Compared to Python, \lang features concise and limited function sets, simplified syntax, and capable as well as predictable logic control. These features enable the \exemode which plays a critical role in reducing the response latency. Additionally, \lang also supports an exception-handling mechanism to handle expected replanning and unexpected errors during the code execution.

\subsection{\textbf{Language Design}}\label{sec:language_design}
In \lang language, the semantics of the system skills and \plan are defined as a \lang program that adheres to the Backus–Naur Form (BNF) syntax specified in \lstref{minispec-syntax}.
% For a low-level skill, the program is a singular statement: a callable function; for a high-level skill, the program is conveyed through multiple statements, including the invocation of other skills and integration of control structures. 
\lang plays a crucial role in achieving efficient and safe \llm usage through four key aspects: (1) Concise and Limited Function Calls: \lang ensures low token count in code generation while preventing unexpected function calls, thereby enhancing efficiency and security. (2) Capable and Predictable Logic Control: \lang supports intricate plan generation and prevents infinite loops, offering versatile and reliable logic control. (3) Simplified Syntax for Streaming Execution: \lang facilitates a simplified syntax set enabling custom parsing and a \textit{Streaming Execution} feature to immediately start code execution whenever an executable piece of the plan is received, reducing response time and improving overall responsiveness. And (4) Exception Handling:  \lang provides mechanisms for handling both expected replanning and unexpected errors during execution, improving planning efficiency and accuracy. Each of these aspects is discussed in detail below.

\subsubsection{Concise and Limited Function Calls}
We design \lang to ensure plans are created with a few numbers of tokens while adhering to a limited set of available functions.
Specifically, each skill has a unique abbreviation \code{skill\_abbr} which is generated automatically based on \code{skill\_name}, e.g., \code{iv} for \code{is\_visible} and \code{tc} for \code{turn\_clockwise}. While executing the plan, only valid callable objects (i.e. system skills and \lang keywords) can be executed, preventing potential safety issues and undefined behaviors.
The usage of skills in \lang follows the syntax of function calls (Listing 1), where the skill abbreviation corresponds to the <function-name> in the function call.
It is important to note that fewer characters do not always result in fewer tokens. For example,  "desc." uses two tokens while "description" uses only one, according to OpenAI's tiktoken tokenizer~\cite{openai2023tiktoken}. However, we found that all two-character words always use a single token. As a result, the abbreviation of each skill is generated to be unique with at most 2 alphabet characters.

\subsubsection{Capable and Predictable Logic Control}
\lang supports conditional control and \textbf{bounded} iteration (a loop block must have a fixed predetermined count), enabling the composition of intricate task plans while guaranteeing that all operations will terminate within a bounded number of steps. 
Furthermore, \lang's syntax supports variable assignment, providing the mechanism for state manipulation. Consequently, while \lang falls short of Turing completeness criteria due to the absence of support for infinite looping constructs, it still possesses substantial computational capabilities.

\subsubsection{Simplified Syntax for \exemode}
The \exemode feature supported by \lang means during the LLM streaming generation, the code execution can start immediately whenever a valid piece of code is received.
We achieve this feature by maintaining a compact and simple syntax, which includes only the logic control, variable assignment, and available system skills, \lang code can be easily parsed by our interpreter. This feature allows \lang to support agile robot control with minimal response time, thereby providing a more engaging and interactive user experience, and being effective for real-time applications. The detail of \exemode will be introduced in~\ref{sec:runtime}.

\subsubsection{Exception Handling}
\label{sec:replan}
\lang supports exception handling with a special keyword \code{replan}, which can be used in 3 ways: (1) \llm can use \code{replan} as a function in the generated plan to improve the planning efficiency and accuracy in a dynamic environment; (2) Users can define custom logic to trigger \code{replan} when certain actions are likely to cause significant scene changes. For example, \replan can be triggered if the drone rotates more than 180 degrees or moves forward over 10 meters, as such actions likely indicate a substantial change in the scene. (3) Whenever a syntax error is found during parsing or a mechanical fault occurs during execution, \code{replan} will be triggered by the interpreter to find a potential fix. 

Once \code{replan} is encountered in the plan, triggered by the interpreter, or thrown by system skills, the execution of the plan immediately stops and signals the \llmc to send a new planning prompt to the \llm to regenerate a new plan.

To further illustrate the use of \code{replan} when the \llm actively uses it in generated plans, consider the following example:
Given the task \task{``Turn around and go to the apple’’} (one used in evaluation~\autoref{sec:eval}), a plan generated without \code{replan} might be 
\begin{lstlisting}[language=Python]
turn_cw(180);goto('apple')
\end{lstlisting}
If there is an obstacle between the drone and the apple, the \code{goto(`apple')} command will fail, causing the task to fail.
With our design, the \llm can first generate a plan:
\begin{lstlisting}[language=Python]
turn_cw(180);replan
\end{lstlisting}
The \code{replan} will then generate new plans to guide the drone around the obstacle and reach the apple based on the updated scene, such as 
\begin{lstlisting}[language=Python]
move_left(50);goto('apple')
\end{lstlisting}

When a task involves an environment that is not fully known at the initial planning stage, generating the entire plan at once with the initial scene description may not be efficient or accurate. 
The \code{replan} function allows the \llm to reason about potential scene changes in a multi-step task, better adapting to environmental changes and avoiding the overhead of generating unnecessary plans in advance compared to single-pass offline planning methods.
Importantly, it is invoked only when necessary, unlike the fully online method where updates are made at every step~\cite{garrett2020online, nageli2017realtime}, striking a good balance between efficiency and usability.

\subsection{System \Skills}
\label{sec:system-skills}

A system \skill is a \lang callable object representing the system's capability. It is typically created by humans and provided to the LLM for use in the plan.
Toward the efficiency goal, we start with the best practices available from the literature such as \cite{liang2023icra} and \cite{vemprala2023arxiv} to design a hierarchical skillset including low-level and high-level skills: low-level skills correspond to callable functions supported by the system, while high-level ones are constructed with other skills and logic controls. All system \skills are blocking and return when their action finishes.
%\autoref{tab:low-level-skills} summaries the designed skills of \system.

%\textit{Python syntax, \lang semantics.}
%\textcolor{blue}{We define the behavior (semantics) of skills using \lang (\S\ref{sec:language_design} and implement these skills using Python syntax due to its user-friendly nature.}
%While we implement skills with Python syntax, we define their behavior (semantics) with \lang (\S\ref{sec:minispec}).  
\autoref{code-skill-example} shows examples of skill implementation in \system. Skill comes with an expressive \code{name} and a \code{desc} (i.e. description), which help the LLM to understand the functionality. 
A low-level skill has the most primitive behavior: a single callable function with arguments as supported by the system, e.g., \lineref{line:callable} of \autoref{code-skill-example}. However, the behavior of high-level skill, as shown in \lineref{line:scan_semantic} of \autoref{code-skill-example}, includes a \lang program as a string. Moreover, unlike low-level skills, a high-level skill does not take any arguments explicitly; instead, it will infer the actual arguments from the positional arguments in its \lang definition. 
%For instance, the \code{\$1} in the \code{scan}'s definition will be inferred as \code{[object\_name: str]} according to the definition of \code{is\_visible}.

\begin{figure}[t]
\begin{lstlisting}[language=Python, label={code-skill-example}, caption={\textmd{\small Examples of \lang skill implementation. A skill has a name, an auto-generated abbreviated name, and a description. Low-level skills have callable and args while high-level ones have \lang definitions.}}, escapechar=|,escapeinside={(*@}{@*)}]
(*@\color{typeflyred}LowLevelSkillItem@*)(
    name="turn_cw", #abbr: tc
    callable=self.drone.turn_cw, (*@\label{line:callable}@*)
    desc="Rotate drone clockwise by certain degrees",
    args=[SkillArg("deg", int)])
    
(*@\color{typeflyred}LowLevelSkillItem@*)(
    name="is_visible", #abbr: iv
    callable=self.vision.is_visible,
    desc="Check the visibility of target object",
    args=[SkillArg("object_name", str)])

(*@\color{typeflyred}HighLevelSkillItem@*)(
    name="scan", #abbr: s
    definition="8{?iv($1)==True{->True}tc(45)}->False", (*@\label{line:scan_semantic}@*)
    desc="Rotate the drone to find a target")
\end{lstlisting}
%\vspace{-1.3em}
\end{figure}

\subsubsection{Low-level skills.} 
\system's low-level skills include fundamental Drone Control, Vision Tool, User Interface, and one special skill. 
Assuming all robotic platforms are equipped with a camera, the only platform-specific skills are Drone Control ones.
The \textit{Drone Control} skills are basic drone maneuvers that are supported by most drone platforms, such as \code{move\_forward} and \code{turn\_cw}.
\textit{Vision Tool} skills offer object-specific details like the target objects' location, dimension, and color, which are supported by popular computer vision tools such as YOLO. 
The \textit{User Interface} skills show text or image feedback to the user.

MiniSpec also features one special skill \spskill, which is particularly important for \system's capability and efficiency by engaging the \llm during the plan execution. 
Calling \spskill in the plan produces a small prompt to the \llm that includes the \code{question} (i.e. the argument of the \spskill) and the latest \scened.
It instructs the LLM to answer the question with minimum words, usually 1 or 2 tokens, and returns with the answer to continue execution.
Such a skill leverages the common knowledge reasoning power of the \llm efficiently at runtime, again, without doing online planning which interleaves planning and taking actions.
For instance, given the task \task{``Go to the edible object behind you''} as shown in Figure~\ref{fig:system-diagrams}, a plan with \spskill can offload the conditional check of \program{``Any edible target here?''} to the \llm and do conditional turns to search for the edible object, as shown in the top of ~\lstref{code-plan-example}. Without using \spskill, the \llm will generate a plan as shown at the bottom of in~\lstref{code-plan-example}, which lists all possible edible objects and searches them individually. 
This is neither efficient for the plan generation since it results in a longer plan nor complete because the list of target objects is finite. Also, in such a case the execution time will be significantly varied depending on the actual object in the scene.
\begin{figure}[t]
\begin{lstlisting}[language=Python]
8{_1=p('Any edible target here?');_1!=True{g(_1);->True}tc(45)}->False
\end{lstlisting}

\begin{lstlisting}[language=Python, label={code-plan-example}, caption={\textmd{\small Example \code{plan}s produced by the LLM (GPT-4) with (top) and without (bottom) \spskill skill for the task "Find something edible". \code{s} and \code{q} are the abbrevations for \code{sweeping} and \spskill, respectively.}}, escapechar=|]
?s('apple')==True{g('apple')->True}
?s('cake')==True{g('cake')->True}
?s('sandwich')==True{g('sandwich')->True}
?s('orange')==True{g('orange')->True}
->False
\end{lstlisting}
%\vspace{-2em}
\end{figure}

% \paragraph{Descriptive Nomenclature}: We intuitively encapsulate the intended behavior in the name and arguments of a \code{skill}. For instance, a \skill enabling the drone to advance a specified distance is aptly denoted as \code{move_forward(distance: int)}. We also provide a supplemental comment in cases where the name might not sufficiently convey the \skill's functionality.

\subsubsection{High-level skills} \system's skillset selectively includes a small number of high-level skills to improve the cost efficiency of \plans and to simplify the programming job of the \llm. 
Low-level skills constitute the entire capability of \system. In theory, the \llm can produce a \plan solely based on them. Such a \plan, however, is verbose and as a result, not cost-efficient. 
On the other extreme, one could provide a large and semantically rich set of high-level skills, converting all previously successful \plans into high-level skills. Because the entire skillset is included in a Planning prompt, a large skillset with numerous high-level skills increases the cost of \llm service.
Moreover, our experience shows that adding too many high-level skills will cause the \llm to produce \plans with only high-level skills and result in incorrect plans. 
% For instance, integrating a skill like \code{find\_food} causes the language model to default to this skill for tasks involving edible items. This approach becomes problematic when a more specific target is needed, such as locating sweet food. In such cases, the language model inappropriately resorts to the generic \code{find\_food} skill, neglecting the need for more precise targeting.
% \lin{Why would LLM ignore ``sweet'' here? I would imagine it will use a loop of ``find\_food and then check if the object is sweet or not''.}
% \gc{gpt-4 will ignore that, even if a task "find something drinkable" will result in a plan with find\_food.}
Balancing between the above two extremes, \system includes a small number of high-level skills. They are \code{scan(obj\_name)}, \code{scan\_abstract(description)}, \code{orienting(obj\_name)}, and \code{goto(obj\_name)}.
%as summarized by \autoref{tab:low-level-skills} (bottom rows). 
These skills are widely used by most tasks to find and interact with target objects, having them in the high-level skillset can simplify and improve the quality of the \plan with little overhead in the planning prompt.

% \lina{Since skills are not introduced yet, the next paragraph needs to move into 5.2}
\lstref{code-example-scan} shows a high-level skill \code{scan} in \lang along with an equivalent Python code representation. \code{scan} rotates the drone for 360 degrees in 8 steps, checking if the given object is present. Whenever the target object is spotted during the loop, the skill returns \code{True}; otherwise, the drone is rotated clockwise by 45 degrees. If the loop completes, the skill returns  \code{False}. In particular, the above skill definition in \lang takes just 18 tokens, while its Python equivalent requires 41 tokens, as measured by the OpenAI tiktoken tokenizer \cite{openai2023tiktoken}.
%More examples can be found in \lstref{prompt-high-level-skills}.
\begin{figure}[t]
\begin{lstlisting}[language=Python]
8{?iv($1)==True{->True}tc(45)}->False
\end{lstlisting}

\begin{lstlisting}[language=Python, label={code-example-scan}, caption={\textmd{\small
Semantic definition of the high-level skill \code{scan} in \lang (18 tokens, top block) and its Python equivalent (41 tokens, bottom block).
}}]
def scan(object_name):
  for i in range(8):
    if vision_skill.is_visible(object_name) == True:
      return True
    drone_skill.turn_ccw(45)
  return False
\end{lstlisting}
%\vspace{-1.5em}
\end{figure}

\subsection{\textbf{\langI}}\label{sec:runtime}

%Interpreting a program in \lang is straightforward due to its small syntax. 
One of the key factors featured by \system to reduce the response time is \exemode, which allows the drone to start action while \llm is still generating the plan. As a comparison, starting execution after the whole plan is received, used by most works, is called Normal Interpreting here. 

\begin{figure}[t]
  \centering
  \includegraphics[width=0.48\textwidth]{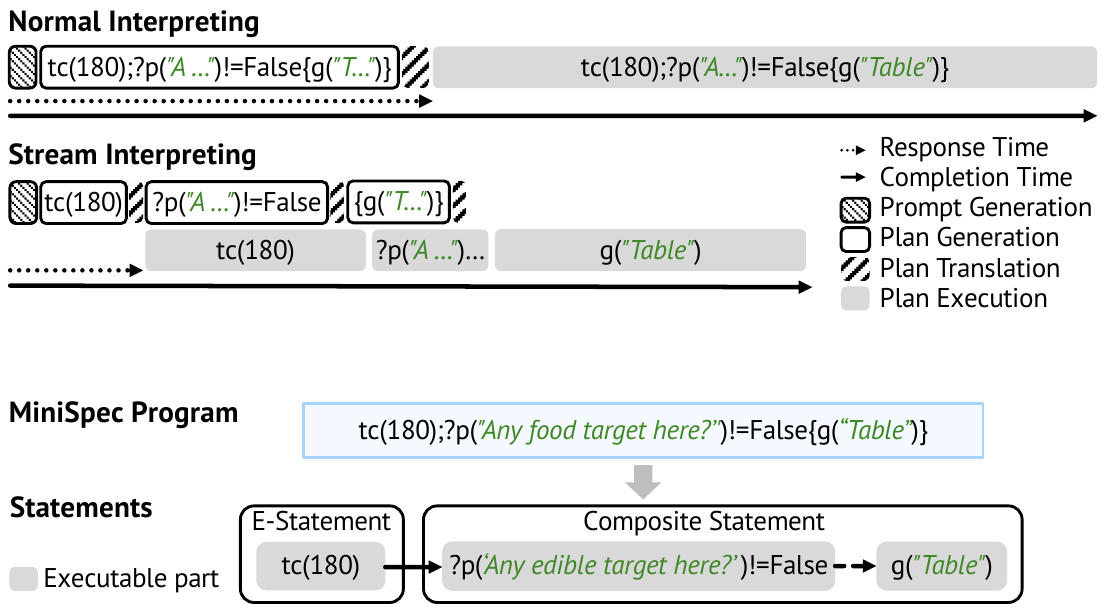}
  %\vspace{-2ex}
  \caption{\textmd{\small Normal vs. Stream Interpreting of a \lang plan and \lang parsing. In Normal Interpreting, \system waits for the whole plan to be received from \llm to start translation and execution. The response time is highly related to the length of the plan and makes the drone less responsive when the plan is long. In \exemode, the response time is reduced to receiving the first executable part of the plan. The bottom half shows the executable part for different types of statements. Note that the network latency is omitted in the figure. }}
  \label{fig:minispec-runtime}
\end{figure}

\paragraph{Stream vs. Normal Interpreting} Below we introduce how \exemode can reduce the response time and improve user experience in real-time applications.
As illustrated in \autoref{fig:minispec-runtime}, with Normal Interpreting, the \llmc feeds the prompt into \llm and the system waits for the whole plan (blue strip) to be received and then starts translation and execution. As a result, the response time varies with the length of plans and can cause the system to be less responsive when instructed to perform complex tasks. In contrast, \exemode treats the plan as a stream, with the design of \lang, our \langI can easily identify executable statements (blue blocks in~\autoref{fig:minispec-runtime} and will be detailed later in this section) and start execution as soon as possible. Such a design reduces the waiting time between the user's command and the robot's action, making the control system more responsive and appropriate for real-time applications. 
A contemporary work, LLMCompiler~\cite{kim2023llm}, employs a similar approach of the prefetching for execution at the task level during LLM decoding to improve latency performance in QA applications, however, the \lang and \langI support this feature at the language level and provides a wider range of applicability.

Next, we introduce how \exemode is achieved through the combination of \lang language design (introduced in~\ref{sec:language_design}) and \lang runtime. 
The \lang runtime is an interpreter that can simultaneously receive, parse, and execute the \lang plan. 
\lang runtime has two threads. One preprocessing thread accepts the streamed LLM response, parsing and formatting the \lang code piece into \code{Statements}. The other worker thread employs an infinite loop to accept \code{Statements} from the first thread through a queue.

A \code{Statement} represents a block of \lang code, which can be a function call, a variable assignment, a conditional block, or a loop block. The former two types are \code{Elementary Statements}, e.g. \code{tc(180)} in~\autoref{fig:minispec-runtime}, and the latter two types are \code{Composite Statements}.
A \lang program is a sequence of \code{Statements}. Whenever an \code{Elementary Statement} is complete or a \code{Composite Statement} is partially complete, it's considered executable, and \lang runtime will send it to the worker thread for execution.

For \code{Composite Statements}, the parsing and execution are various depending on the type. A condition \code{Statement} starts with a condition (with a leading `?'), which can be a combination of intersection and union of sub-conditions. Each sub-condition can involve the comparison between the return value of a skill call and literal values. Once the condition is completely received, the \code{Statement} will be sent to the worker thread without waiting for the body block (embraced by ``\{ \}''). While the worker thread is executing the condition, the other thread will continue to append sub-statements into the body block.
If the condition results is \code{True}, the worker thread will start executing the body block. In rare cases, the body block is still empty after the condition execution. In these cases, the worker thread will wait until the next executable \code{Statement} to be appended into the block body.

For \code{Elementary Statements}, the process is straightforward. After a complete one is received by the preprocessing thread, it will be sent to the worker thread for execution. The worker thread simply calls the system skill (or does variable assignment) in blocking mode.

Notably, in our application, most \code{Statements} take significantly longer time to execute than the time to be generated by LLM and cost by preprocessing.

\begin{figure*}[th]
  \centering
\begin{subfigure}{0.26\textwidth}
    \centering
    \includegraphics[width=\textwidth]{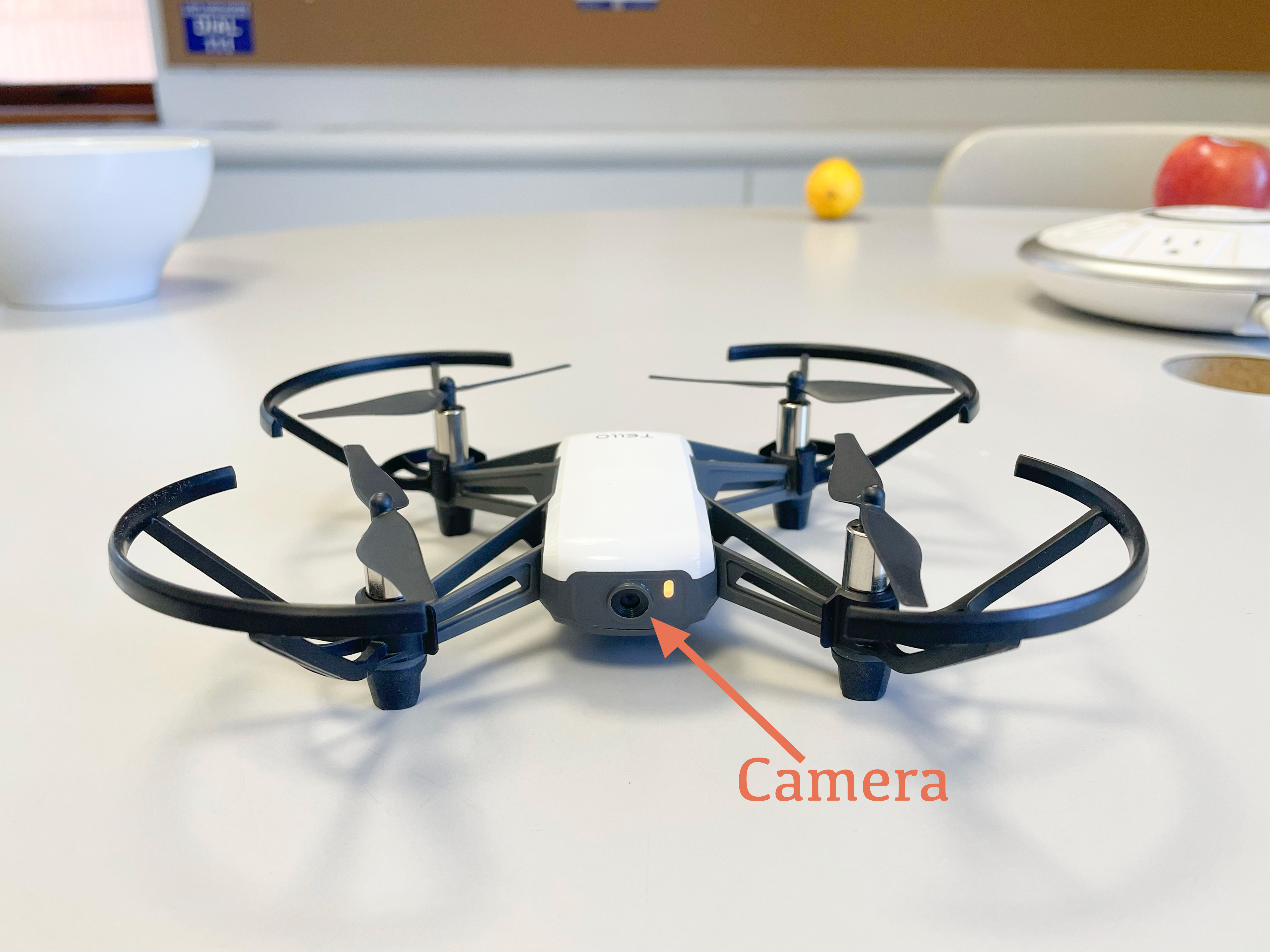}
    \caption{\textmd{\small Tello drone with a front camera}}
    \label{fig:setup-drone}
  \end{subfigure}
  \begin{subfigure}{0.265\textwidth}
    \centering
    \includegraphics[width=\textwidth]{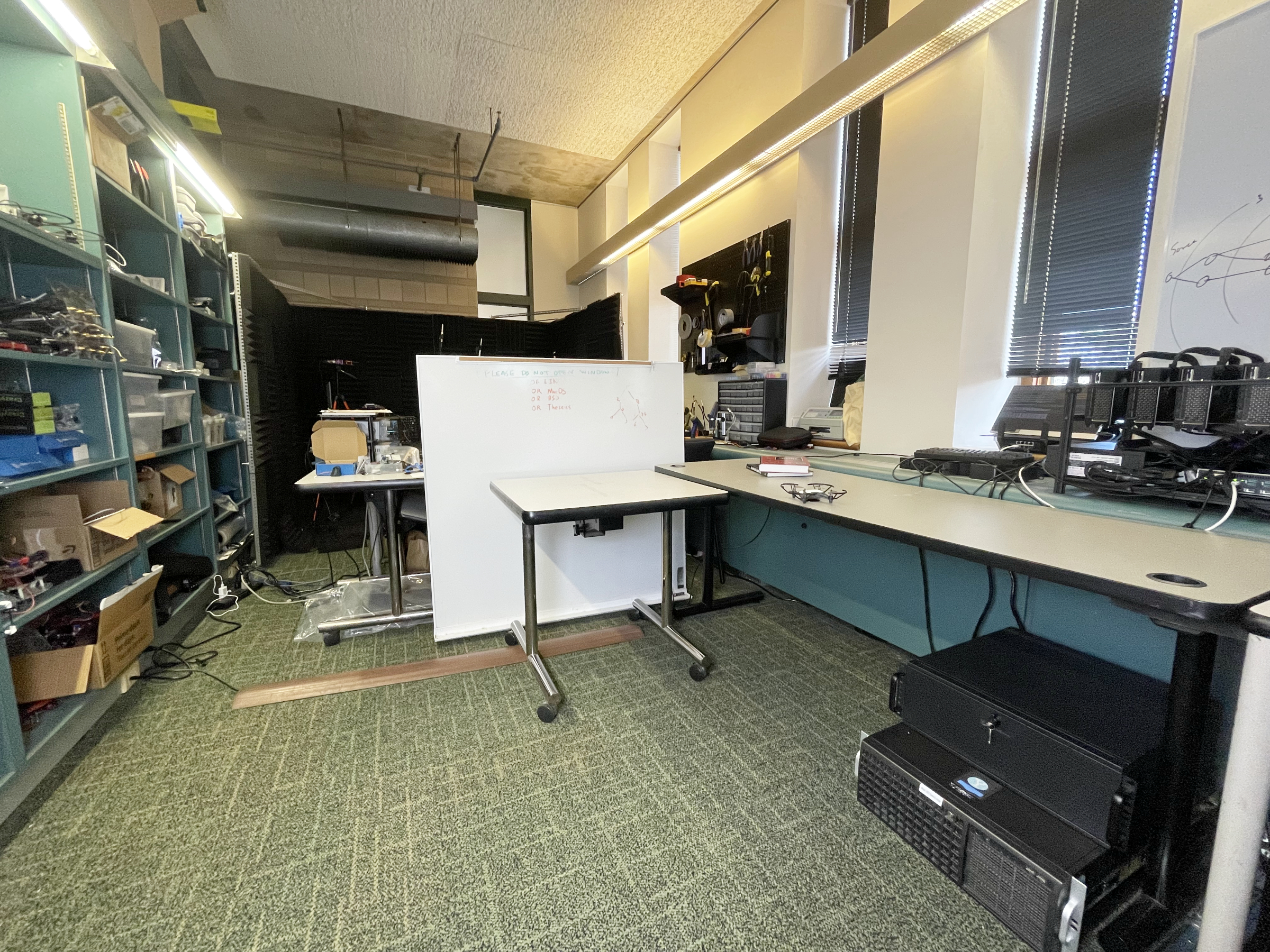}
    \caption{\textmd{\small Evaluation environment setup}}
    \label{fig:setup-environment}
  \end{subfigure}
      % \hfill 
  % \hfill 
  \begin{subfigure}{0.265\textwidth}
    \centering
    \includegraphics[width=\textwidth]{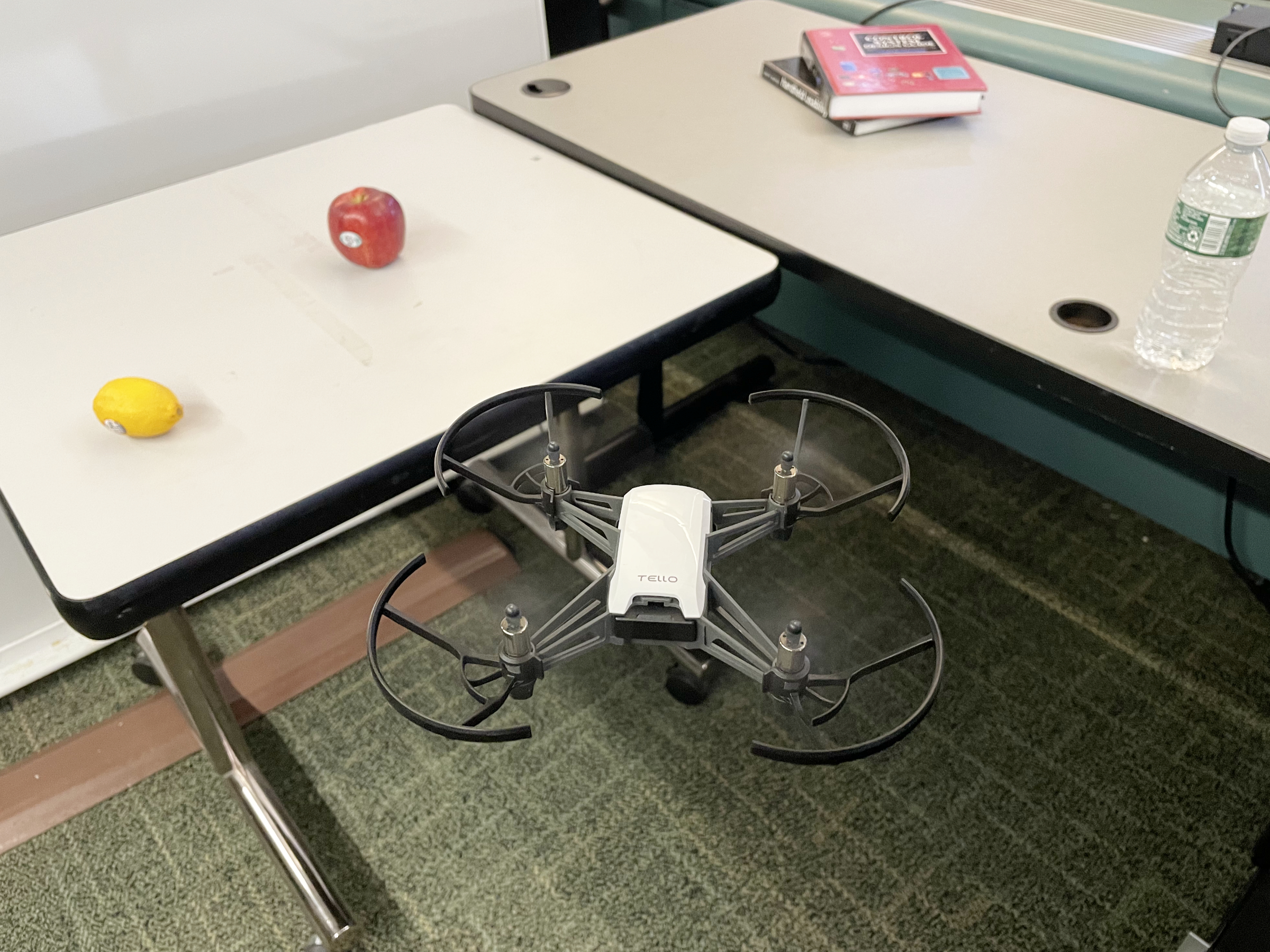}
    \caption{\textmd{\small Task 2 execution}}
    \label{fig:setup-execution}
  \end{subfigure}
  \begin{subfigure}{0.19\textwidth}
    \centering
    \includegraphics[width=\textwidth]{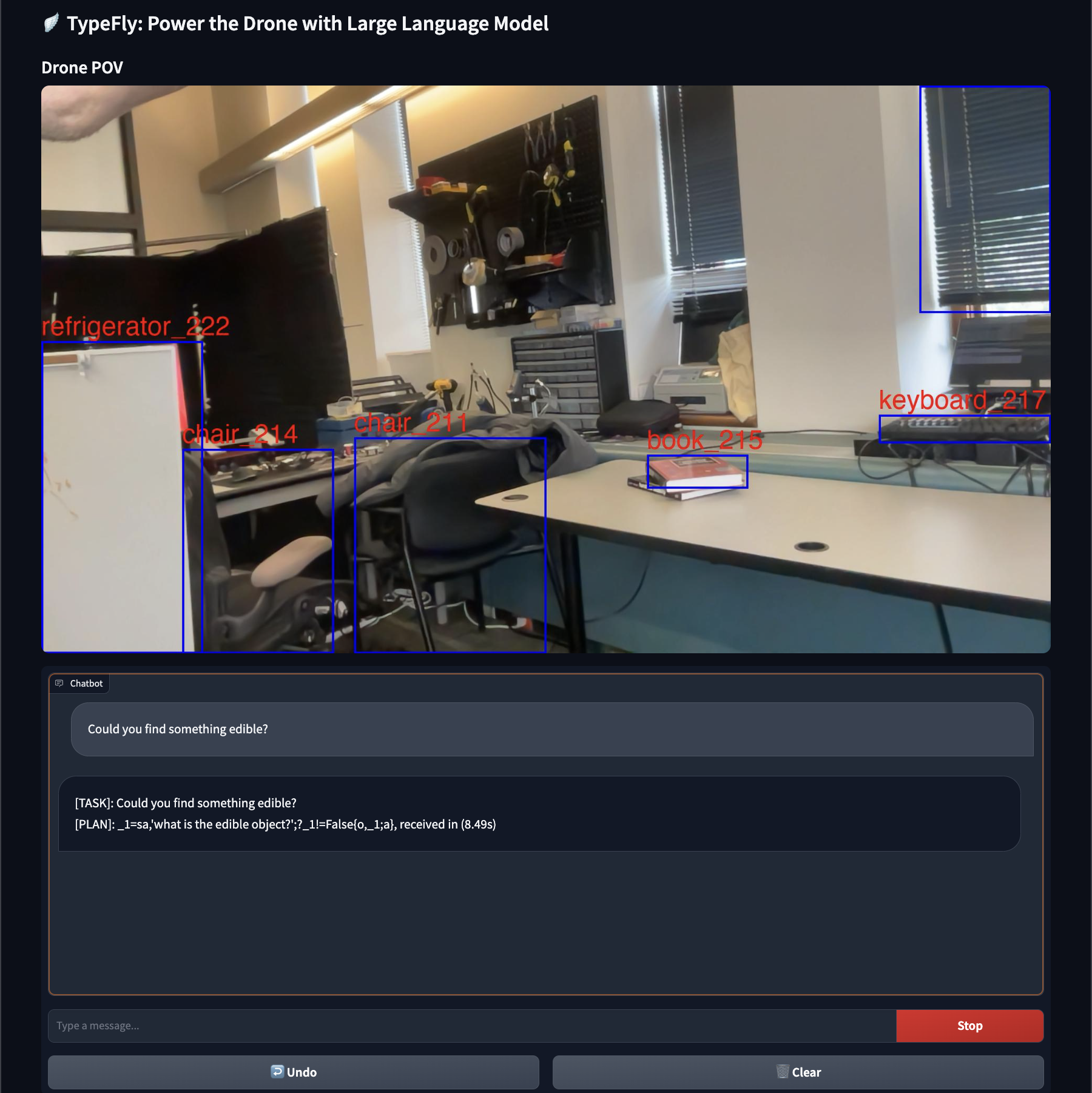}
    \caption{\textmd{\small \system user interface}}
    \label{fig:setup-interface}
  \end{subfigure}
  % \vspace{-1em}
  \caption{\textmd{\small Evaluation setup and the screenshot of \system interface. We use a cheap off-the-shelf drone with video streaming and programmable control API in our evaluation, showing the potential of \system's portability for other kinds of robots. The experiments are done inside a typical office area without any external infrastructure except for our edge server and WiFi.}}
  \label{fig:setup}
\end{figure*}

\section{Implementation}
\label{sec:implementation}
%We next report the prototype hardware and software implementation of \system below.
We next provide the necessary information about our implementation of \system on which the evaluation is based. 

\noindent
\textbf{Hardware.}
\system uses a commercial-off-the-shelf drone and an in-house edge server as the hardware deployment. The drone, Tello, is made by Ryze Tech (\autoref{fig:setup-drone}). It offers Wi-Fi video streaming (at most HD720P/30 frames per second (FPS)) and user-friendly operations through a smartphone application. Moreover, it comes with a Python interface library which provides APIs for both control and video streaming.
% In particular, the Tello drone APIs used in \system are \code{takeoff}, \code{land}, \code{streamon}, \code{streamoff}, \code{up}, \code{down}, \code{left}, \code{right}, \code{forward}, \code{back}, \code{cw}, and \code{ccw}. Their more detailed definitions can be found in Tello SDK User Guide \cite{dji2023tellosdk}. These APIs are all basic drone/camera controls and can be easily supported by other drone platforms such as Parrot \cite{parrot2023drones} and XYZ.
Our edge server is equipped with a Nvidia RTX4090 GPU, a 16-core Ryzen 7950x CPU, and 64 GB RAM. All system-related software, with the exception of the \llm (OpenAI GPT4), operates on this server and collectively consumes at most $20\%$ of the total available resource.

\noindent
\textbf{Software.}
The \system software comprises approximately 2500 lines of Python code and is structured as follows: (1) A User Interaction Module: \system users interact with the system via a web-based interface, as shown in \autoref{fig:setup-interface}. Through this interface, the user can provide task descriptions, observe real-time video feedback from the drone, and read \sceneds output by the \ve. (2) \ve: The \ve generates the scene description based on the drone-captured image. Here we opt for YOLOv8 due to its rapid processing speed and its ability to provide precise geometric information, such as the location of objects.
(3) \llmc: This part integrates the user's task description with the system description and the current scene description from the \ve. Then it composes a prompt and sends it to \llm (OpenAI GPT4 API) for plan generation. (4) \langI: Detailed in~\ref{sec:runtime}, the \langI functions to interpret and execute the \lang \plan generated by the \llm.
%Our implementation can serve up to 5 drones concurrently on the edge server, due to the limit of the GPU: five \ve (YOLOv8 in our case) instances will fully utilize the RTX4090 in the edge server. 
%Our system can easily scale up to support more drones with more GPUs or more powerful GPUs. 
% \lin{Lin revisit: forward reference to opportunities of optimization.}
% \gc{or we can make both video stream and ve processing sync at 20 FPS, it won't affect the overall system's ability to fulfill a task}

Here we provide a more detailed design overview of \llmc and \ve.

\textit{\llmc.}
We opt for a long-prompt short-output strategy instead of generating rich output (e.g. Chain-of-Thought~\cite{wang2023arxiv} or ReAct~\cite{yao2023react}) for the use of \llm for two reasons: (1) With our goal of low latency control, we expect the LLM to generate only the necessary tokens to minimize the plan generation time. (2) Given the first reason, in order to improve the planning accuracy, we use few-shot prompting with detailed explanations and guides.
The drawback of this design is that, with the current cost model of LLMs, \system would result in a higher cost than a system that uses a smaller prompt. However, we observe that works like \cite{gim2023prompt} have shown the potential of reducing the cost for LLMs service of a big structured prompt by reusing the prompt cache. We believe that the cost of \system will be reduced in the future.

Specifically, the prompt generator will construct a comprehensive prompt that incorporates different essential elements for different purposes. (1) For task planning, the prompt integrates the user's task description, the current scene information from \ve, a detailed description of the language's skill set and syntax, and a list of few-shot examples for demonstrating how \lang can be used to formulate executable solutions that meet the task objectives. Moreover, we provide some natural language guidance that aids the LLM in how to use the system skills, how to reason for the task, and the requirements for the output. 
For example, we provide insight into the proper use of \spskill: only when the required information is missing in the current scene, \spskill should be called with an appropriate question.
(2) For exception handling (replan), the prompt includes the basic planning prompt introduced above and the history of the previous plan, enabling the LLM to determine the extent of plan execution and generate a subsequent plan accordingly. (3) For \spskill, we opt for a lightweight question and answer (Q\&A) prompt including basic rules and two examples for response generation.

Notably, the system skills in the prompt are robot-specific. \llmc defines the low-level skills required for a drone with the \code{DroneWrapper} abstract class, which can easily be applied to other drone platforms with a Python interface for basic movement control and video streaming. When working with other types of robots, developers only need to design the skillset for the robot and update the planning prompt accordingly.

\textit{\ve.}
Our \ve uses YOLOv8 and it generates a list of detectable objects with their name of types and bounding boxes. Object type is used for task reasoning with \llm and the bounding box is necessary for locating and approaching the target object. These kinds of information are essential when privacy is a concern, as it is important to avoid sending images to the \llm in such cases. The YOLOv8 service is deployed independently to the other part of the system following a micro-service-like architecture and it can be accessed through \code{gRPC}~\cite{google2023grpc} request. 
While the current system exclusively uses YOLO \cite{jocher2023yolo} object detection, such an architecture design provides a flexible foundation, facilitating seamless expansion to incorporate additional vision services, e.g. CLIP~\cite{radford2021arxiv}, BLIP~\cite{li2022blip}, Yolo-World~\cite{cheng2024yolow}, in future iterations.

\begin{table*}[th]
  \caption{\textmd{\smaller Benchmark tasks used in the reported evaluation. We define 5 types of tasks to test \system's capabilities thoroughly. Tasks 1-3 test \system's basic ability to reason and plan; Tasks 4-6 test using \llm in execution to generate offline plans for the undetermined targets; Tasks 7-8 test several more complex planning scenarios; Tasks 9-10 test the \system's ability of exception handling; Last task evaluates the safety of \lang to generate a plan with termination.}   
  }
  \label{tab:task_list}
  % \vspace{-1em}
  \footnotesize
  \begin{tabular}{|m{1.3cm}|m{0.3cm}|m{9cm}|m{5.5cm}|}
    \hline
    \textbf{Categories} & \textbf{ID} & \textbf{Task Description} & \textbf{Scene Setup}\\
     \hline
     \hline
\multirow{3}{1.3cm}{Basic Planning}
& 1 & Go and take a picture of the chair. & a chair in sight \\\cline{2-4}
    & 2 & Could you find an apple? If so, go to it. & an apple in the scene \\\cline{2-4}
& 3 & Go to the largest item you can see right now. & a person, apple, and keyboard in sight \\\hline
\multirow{3}{1.3cm}{LLM in Execution}
& 4 &  Find something yellow and sweet. & banana and lemon on the table behind the drone \\\cline{2-4}
& 5 & Can you find something for cutting paper on the table? The table is on your left. & table on the left with a pair of scissors \\\cline{2-4}
& 6 & Find a chair and go to the object that is closest to the chair. & a chair on the back with an apple on the chair and a bottle behind the chair \\\hline
\multirow{2}{1.3cm}{Complex Planning}
& 7 & Move up for 1m and check the top of the cabinet, if you see anything red and sweet, take a picture of it. Otherwise, return to the original position. & an apple on top of the cabinet \\\cline{2-4}
& 8 & Can you find something for me to eat? If you can, go for it and return. Otherwise, find and go to something drinkable. & only coke on the left table without any other food \\\hline
\multirow{3}{1.3cm}{Incremental Planning \& Replanning}
& 9 & Turn around and go to the apple. & an apple on the table behind the drone with a chair blocking in between \\\cline{2-4}
& 10 & If you can see more than two people behind you, then turn to the tallest one that is behind you. & 3 people in sight and 2 other people in the back of the drone \\\hline
Safety of \lang
& 11 &  Turn around in a 45-degree step until you see a person with a cup in hand. & none \\
    \hline
  \end{tabular}

\end{table*}

\section{Evaluation}
\label{sec:eval}

% \lina{Keep the existing evaluations as part of Ablation Study (Typefly without incremental programming. This means Figure 5 should be updated to add a new series (with IP). Table 3 should be updated: either replacing the data with the new Typely, or adding a new table with the data with new Typefly}

We evaluate \system with a set of increasingly challenging tasks as summarized by \autoref{tab:task_list}. We seek to answer the following questions:
\begin{itemize}[leftmargin=1em]
\item What kinds of tasks are feasible vs. infeasible for \system? 
\item When and why would \system fail?
\item How effective is \lang and \langI in improving \system's efficiency and user experience?
\item How effective is \spskill in improving \system's efficiency and capability?
\item How effective is \replan in improving \system's capability?
\end{itemize}
% \lina{We need to preemptively defend why we did not compare Chatfly with existing systems. Rather, we reported ablation studies of Chatfly itself. I think the reasons are: (1) many existing systems do not have an implementation; those that have implementations are not open-source; (2) The ablation study is fair to reveal the impact of MiniSpec vs Python?.....}

We choose to evaluate the effectiveness of \system's key innovations via ablation studies, instead of a direct comparison to related systems, for the following reasons. First, related systems use various robotic platforms (and simulators) in their evaluations. While their software may be open-source, their robotic hardware is not generally available, which makes an apple-to-apple comparison difficult, if possible at all.  This is particularly true because system skills and prompt engineering are highly specific to the robotic hardware.
Second, the key innovations of \system, i.e., \lang and \langI, are agnostic of the robotic hardware and as a result, their impact can be revealed by proper ablation studies.

\subsection{Setup}
\label{sec:setup}

\noindent
\textbf{Physical environment.} The reported evaluation results were obtained from flying the drone inside an office room, as shown in~\autoref{fig:setup-environment}. The room includes tables, chairs, and a variety of objects. Among these objects, some are directly related to the tasks while others are not.
During evaluation, we found, not surprisingly, that a proper lighting condition is crucial to the performance of YOLO and the drone's stability. 
%\gc{drone's stability relies on optical flow sensor which is sensitive to lighting}
% During evaluation, we found, not surprisingly, that a proper lighting condition is crucial to the performance of YOLO, the key vision encoder in our system, and the drone's stability.
%We observed that YOLO struggles to accurately detect target objects in environments with insufficient brightness or when the color temperature is excessively warm. Additionally, the drone also relies on the proper light and texture of the carpet to hover stably.
Our setup has a standard office building lighting condition with approximately 500 lux brightness and a neutral color temperature in the range of 4000K to 5000K, according to our measurement.

\noindent
\textbf{Benchmark tasks.}
During the evaluation, we tested \system with a list of tasks to test its limit and effectiveness: when it works, when it does not, how efficient the system is?
Due to the space limit, we must curate the benchmark tasks such that they demonstrate both the capability and the limit of \system, illuminate the roles of key ideas of \system, such as \lang, \spskill, and \replan, as well as point the direction for future improvement. Guided by these objectives, we select 11 tasks of five categories as summarized by \autoref{tab:task_list}. 
Importantly, we define a task by task description and scene setup, because performance depends on both.
Our benchmark tasks are substantially more complicated than those used by prior work with drones such as~\cite{vemprala2023arxiv,tazir2023wcse}; none of them would be supported by prior systems. 

\noindent
\textbf{Metrics.} Given a task, we evaluate the performance of \system with the following metrics.

\begin{itemize}[leftmargin=1em]
 \item whether \system accomplishes the task, i.e., success or not. This is about the semantic correctness of the \plan produced by \prog{GPT4} and the feasibility of the task with \system.
\item the response time it takes \system to generate the initial plan (R-Time). The response time means the duration from the user query to the first action of the drone.
\item the task completion time it takes \system to finish the task (C-time). The total time refers to the duration from the user query to the task completion by the drone.
\item the total number of tokens in the output \plans (Token \#)
\end{itemize}

\subsection{Overall performance of \system}
\textbf{Success rate.}\label{sec:success-rate}
\autoref{tab:task_list_evaluation} presents the metrics of all benchmark tasks. 
We repeated each task 10 times and reported the average along with the standard deviation (Std).
\system succeeded in finishing most tasks, except for several failure cases for \textit{Task 9: Turn around and go to the apple}.
% \textbf{Failure Cases}
% \label{sec:failure-cases}
% We next take a deep dive into the failure cases of tasks 10 and 11.
For this task, we aim to test \system's capability to generate a correct \plan when the engaged scene setup changes. The scene setup includes an apple on the table behind the drone with a chair blocking in between. \system fails to accomplish this task in 3 out of 10 runs. The failure is rooted in that \system cannot well estimate the geometric size of the obstacle (the chair). It falsely assumes the apple is reachable after the drone moves leftwards a little bit. However, the edge of the chair can still block the drone from reaching the apple.

\noindent
\textbf{Plan length and Response time.} The length of the generated plan for the benchmark tasks vary from 6 (task 3) to 47 (task 8). It is obvious that the length is related to the task complexity. The response time includes the network round trip time (RTT) to \prog{GPT4} and the latency of \prog{GPT4} generation. Because the network RTT is on the level of 10s milliseconds, the response time is dominated by GPT4's inference latency. Not surprisingly, in the Normal Interpreting mode, the response time is highly correlated with the \plan length: generally, shorter plans experience shorter response time. However, with \exemode, the response time is similar for all tasks due to the design of \lang and \langI, which highlights our efficient target.

\noindent
\textbf{Task completion time.} The task completion time of a \plan is also roughly related to its token number. The most complex task (task 8) takes the longest time to finish. Environmental variation and the possible use of a bounded loop reduce the correlation between \plan length and task completion time.

% \textit{Plan length and Response time.} The length of the generated plan for the benchmark tasks vary from 6 (task 3) to 47 (task 8). It is obvious that the length is related to the task complexity. The response time includes the network round trip time (RTT) to \prog{GPT4} and the latency of \prog{GPT4} generation. Because the network RTT is on the level of 10s milliseconds, the response time is dominated by GPT4's inference latency. Not surprisingly, in the Normal Interpreting mode, the response time is highly correlated with the \plan length: generally, shorter plans experience shorter response time. However, with \exemode, the response time is similar for all tasks due to the design of \lang and \langI, which highlights our efficient target.

% \textit{Task completion time.} The task completion time of a \plan is also roughly related to its token number. The most complex task (task 8) takes the longest time to finish. Environmental variation and the possible use of a bounded loop reduce the correlation between \plan length and task completion time.

\begin{table}[t]
  \caption{\textmd{\small Evaluation results of the benchmark tasks based on 10 runs of each. Most tasks are successfully accomplished by \system. The average response time is relatively independent of the length of the plan due to our \exemode design. Task 9 has several failure cases which will be discussed in \S\ref{sec:success-rate}.}}
  \label{tab:task_list_evaluation}
  \footnotesize
  \begin{tabular}{|m{0.4cm}<{\centering}|m{1cm}<{\centering}|m{1.6cm}<{\centering}|m{1.6cm}<{\centering}|m{1.6cm}<{\centering}|m{1.6cm}<{\centering}|}
    \hline
  \multirow{2}{*}{\textbf{ID}} & \multirow{2}{*}{\textbf{Success}}& \multicolumn{3}{c|}{\textbf{Ave./Std.}} \\ \cline{3-5}
     &   & \textbf{  R-Time (s)} & \textbf{ C-Time (s)} & \textbf{ O. Token \#}\\
     \hline
     \hline
1  & 10/10 & 1.24/0.13 & 7.83/0.85 & 7/0  \\
2  & 10/10 & 1.28/0.08 & 12.51/1.29 & 10/0  \\
3  & 10/10 & 1.26/0.10 & 7.28/0.98 & 6/0  \\
4  & 10/10 & 1.35/0.15 & 15.84/2.37 & 21/0  \\
5  & 10/10 & 1.19/0.10 & 11.07/1.37 & 27/0  \\
6  & 10/10 & 1.22/0.12 & 9.95/1.00 & 38/0  \\
7  & 10/10 & 1.19/0.13 & 38.00/2.56 & 40/0  \\
8  & 10/10 & 1.32/0.11 & 47.90/5.86 & 47.1/0.6  \\
9 & \textbf{7/10}  & 1.10/0.13 & 15.44/1.90 & 15/0  \\
10 & 10/10 & 1.22/0.13 & 12.35/1.41 & 35.0/6.3  \\
11 & 10/10 & 1.60/0.13 & 26.10/3.19 & 33/0  \\
    \hline
  \end{tabular}
%  \vspace{-1em}
\end{table}

\begin{figure}[t]
  \centering
   \includegraphics[width=0.48\textwidth]{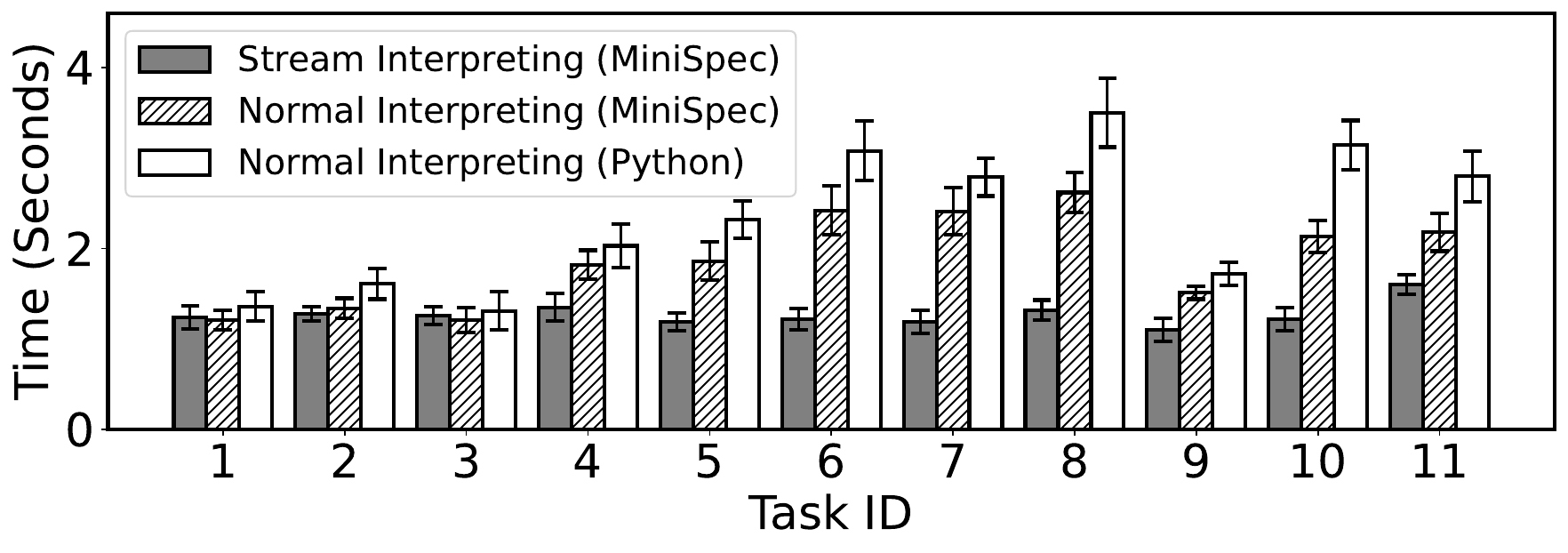}
  \caption{\textmd{\smaller Response time comparison of the system using \lang with \exemode, \lang with Normal Interpreting, and Python with Normal Interpreting. The Python plan adheres to the same logic as the \lang plan. Using \lang results in at most $32\%$ reduction in response time and further employ \exemode can achieve up to $62\%$ response time reduction as well as provide a more consistent performance when compared with the Python baseline.}}
  \label{fig:response-time-compare}
\end{figure}

\begin{figure}[t]
  \centering
   \includegraphics[width=0.48\textwidth]{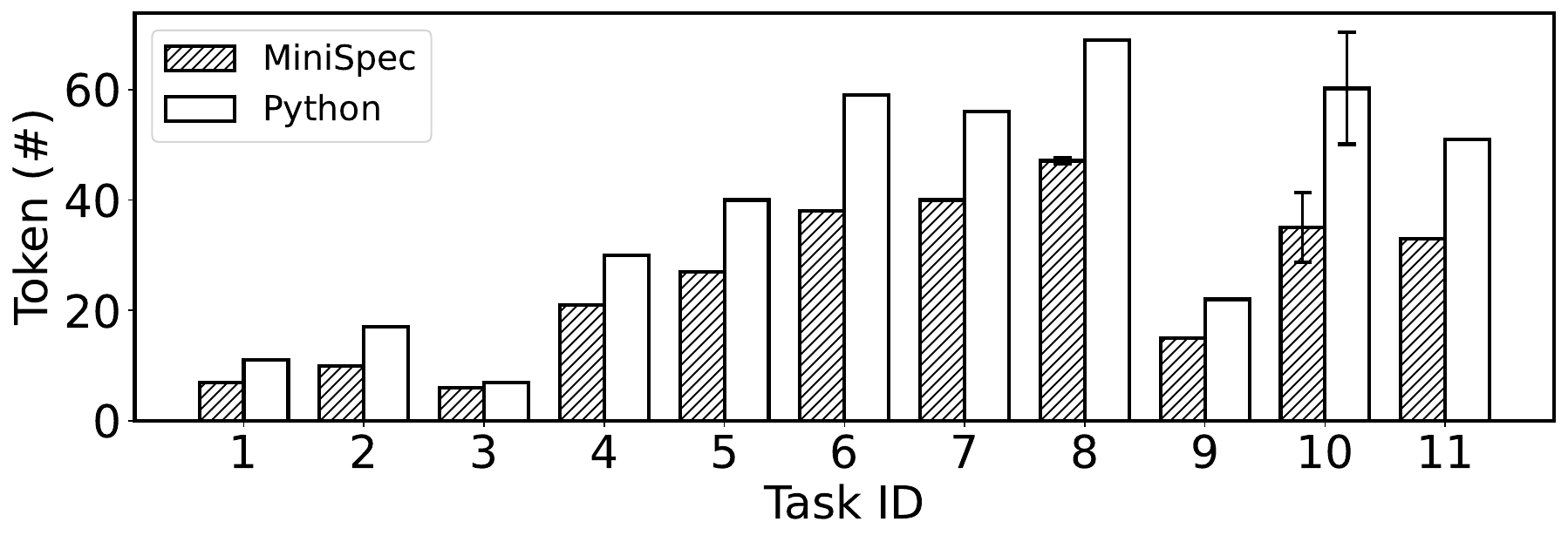}
  \caption{\textmd{\smaller Token number comparison of the system using MiniSpec vs. using Python to generate the plans. The Python implementation also features prompt engineering and system skill design. Using MiniSpec results in up to a $42\%$ reduction in token numbers and an average reduction of $34\%$.}}
  \label{fig:minispec-python-compare}
\end{figure}

% We considered several ways to fix this problem, which were not included in the rest of the evaluation. The first way is to modify Guide 4 from \autoref{prompt-rules} to remind \prog{GPT4} to focus on the new \scened after drone movement. 
% In general, this solution is not completely reliable: occasionally, \prog{GPT4} would still fail to understand the scene has changed after drone movement.
% A more reliable but less scalable solution is to include one example in the Planning prompt as below. Our experience showed that this example is surprisingly effective and general. With it, \prog{GPT4} seems to ``understand'' that the set of visible objects will change after drone movement, for any type of object, despite that the example is only about chairs.

% \begin{lstlisting}[label={lst:additional-example}, escapeinside={(*@}{@*)}]
% (*@\color{typeflyred} Example 9@*) (not used in reported evaluation results):
% scene: [chair_13, chair_2, laptop_2], task: [A] Go to the chair that is behind you.
% response:tc,180;o,chair;a
% reason: chair_13 and chair_2 are not the target because we want the one behind you. So we turn 180 degrees then orient and approach the object chair.
% \end{lstlisting}
% \vspace{-0.2em}

\subsection{Effectiveness of \system component}

\subsubsection{Effectiveness of \lang}
\lang serves as the key to improving task planning efficiency by minimizing both the response time and the number of output tokens required for a given plan. As illustrated in \autoref{fig:response-time-compare}, the \lang itself with Normal Interpreting achieves at most a $32\%$ reduction in response time. Moreover, with \exemode, \lang can save up to $62\%$ response time, thereby significantly improving the robot's responsiveness in real-time and interactive applications. This reduction in response time also leads to more consistent performance across tasks of varying complexity, ultimately enhancing the user experience.
In \autoref{fig:minispec-python-compare}, we also show the complete plan token numbers for the whole task list when using \lang and Python. On average, \lang reduces the output token number by 34\%. 

\subsubsection{Effectiveness of \prog{probe}}
\label{sec:eval-probe}
The \spskill plays a significant role in improving the system's efficiency,  both \plan generation and execution. We evaluated the absence of \spskill on tasks 4, 5, 6, 7, and 8, since they are related to abstract object identification in unseen environments. For tasks 4, 6, and 7, our system fails to generate a feasible plan without \spskill since the planner cannot determine the target object at the planning stage. For task 5, remote LLM happens to generate a plan to find ``scissors'' which is a reasonable target for cutting paper, however, this is not reliable. For task 8, the generated plan loops over several edible and drinkable targets, making the plan very long and inefficient.

\begin{figure}[t]
  \centering
    % \hspace{3em}
   \includegraphics[width=0.48\textwidth]{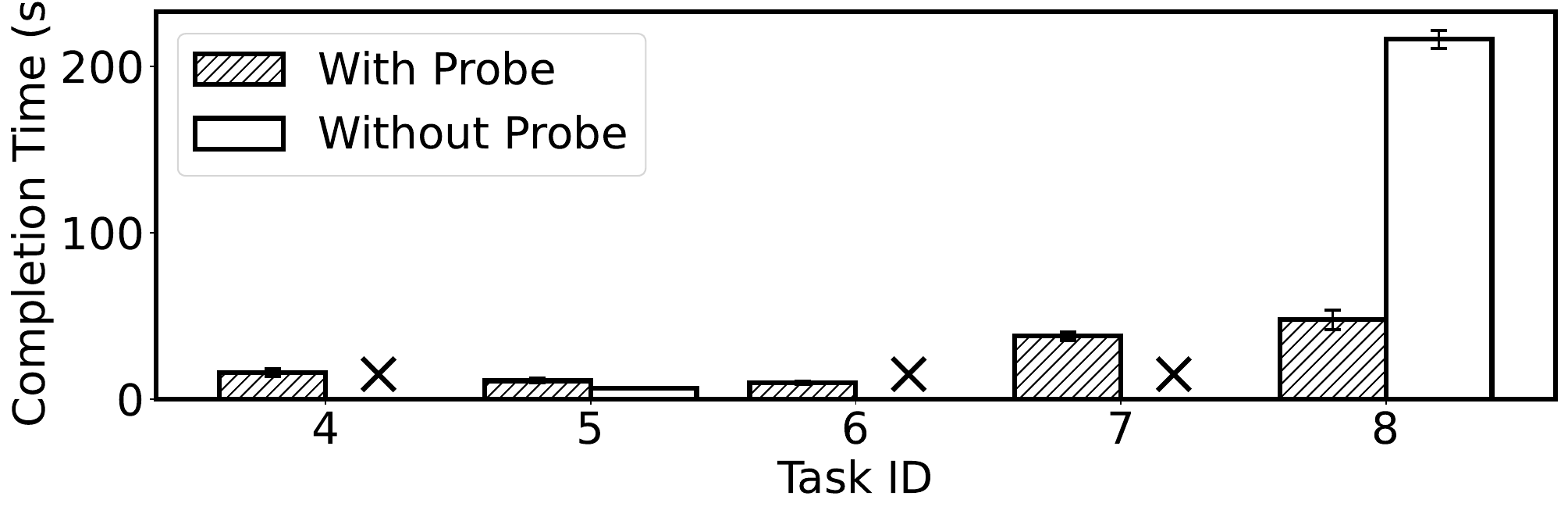}
   %\vspace{-1em}
  \caption{\textmd{\smaller Task Completion time of the plan generated by the system with/without the skill \spskill for the five tasks. For tasks 4, 6, and 7 without \spskill, \system fails to fulfill the task. In task 5, the system works coincidentally by directly searching for scissors since it's a reasonable target for cutting paper. Without \spskill, task 8 results in a long plan and long completion time with enumeration for every edible and drinkable object.}}
  \label{fig:probe-compare}
\end{figure}

In \autoref{fig:probe-compare}, the advantages of integrating \llm during execution are evident. By doing so, there is a significant reduction in output token length, plan request latency, and overall task completion time. This is because when equipped with the \spskill function, \llmc can apply the knowledge base of the \llm to a dynamic environment more efficiently with offline planning.

\subsubsection{Effectiveness of Exception Handling}
\label{sec:eval_incremental}
We conduct a similar evaluation for tasks 9 and 10 without \replan to evaluate the effectiveness of our exception handling. The results reported in \autoref{tab:task_replan_evaluation} show that \replan is crucial in improving the success rate and efficiency for tasks related to the unseen environment. Below we dive into each task and reveal how \replan helps.

\textit{Task 9}:~~ As we discussed before, even with scene-based incremental planning, \system fails to accomplish this task in 3 out of 10 runs due to the lack of geometric understanding of the environment. Without \replan, \system is unable to know the obstacle (the chair) at the initial planning stage and fails to accomplish the task in all 10 runs.

\textit{Task 10}\label{sec:task10}:~~ This task is conducted in a room with 3 people in sight and 2 other people in the back of the drone. \system fails to accomplish this task in 2 out of 10 runs. The failure is rooted in that \llm does not understand the potential scene changes when actions like ``turn around'' are not explicitly mentioned in the task description. It falsely assumes the 3 people in the original \scened are the same set of visible people after the drone turns 180 degrees. Indeed, if we modify the scene setup such that there is nobody in the original \scened, \system will produce a correct \plan. Since task 10 is all about people originally behind the drone, whether there were people originally before the drone should not have mattered. 
With \replan, after the drone turns 180 degrees, \system can generate a correct \plan by re-counting the people and determining the tallest one from the new \scened.
\begin{table}[t]
  \caption{\textmd{\small Evaluation of tasks 9 and 10 with/without \replan. For task 9, without knowing the potential obstacles the task fails all the runs; For task 10, the \llm occasionally confuses the foreground people with the actual targets and mistakenly believes that the condition is met. The task will fail without \replan when such a mistake happens.}}
  \label{tab:task_replan_evaluation}
  \footnotesize
  \begin{tabular}{|m{0.4cm}<{\centering}|m{1cm}<{\centering}|m{1.6cm}<{\centering}|m{1.6cm}<{\centering}|m{1.6cm}<{\centering}|m{1.6cm}<{\centering}|}
    \hline
      \multirow{2}{*}{\textbf{ID}} & \multirow{2}{*}{\textbf{Success}}& \multicolumn{3}{c|}{\textbf{Ave./Std.}} \\ \cline{3-5}
     &   & \textbf{  R-Time (s)} & \textbf{ C-Time (s)} & \textbf{ O. Token \#}\\
     \hline
     \hline
     \multicolumn{5}{|c|}{With \code{replan}}\\\hline
9 & \textbf{7/10}  & 1.10/0.13 & 15.44/1.90 & 15/0 \\
10 & \textbf{10/10}  & 1.22/0.13 & 12.35/1.41 & 35.0/6.3 \\\hline\hline
\multicolumn{5}{|c|}{Without \code{replan}}\\\hline
% 9 & \textbf{0/10}  & 1.1/0.2 & 14.0/0.8 & 8.0/0 & 2281.3/3.2 \\
9 & \textbf{0/10}  & - & - & - \\
10 & \textbf{8/10}  & 1.19/0.21 & 13.82/1.53 & 27.2/0.8 \\
    \hline
  \end{tabular}
  % \vspace{-1em}
\end{table}
\section{Concluding remarks}
\label{sec:concluding}

In this paper, we introduce \system, a system that allows drones to accomplish low-latency drone control with natural language commands. \system achieves this through a synthesis of edge-based vision intelligence, innovative custom programming language \lang, and prompt engineering with LLMs. Through a series of progressively complex drone tasks, \system demonstrates its ability to substantially reduce the system's response time and provide a more consistent user experience with efficient task completion. Crucially, the integration of \spskill and the exception handling mechanism (\replan) of \lang in \system markedly enhances the chance of accomplishing complicated tasks.
Our experience with \system also reveals numerous challenges to tackle as well opportunities to explore in future work, which we will briefly discuss below.

The tasks involved in our evaluation can mostly be accomplished through a simple scene description based on the derived object list and their positions.
However, for more complex tasks that require a comprehensive understanding of the scene and environment, \system falls short without historical observation memory and environment modeling.
While past works have utilized sensors like Lidar, we aim to explore the drone’s spatial modeling and understanding in the future, particularly on lightweight, cost-effective drones without many high-precision sensors.

The interaction method in \system involves sending the complete prompt content to the LLM each time. For different tasks, only the scene description and task description can vary, comprising less than 8\% of the total prompt length. Recomputing the repeated content wastes resources and increases latency.
Recent approaches, such as using cache to reuse attention states, have effectively reduced latency, especially time-to-first-token~\cite{gim2023prompt}. We will explore this further to improve the speed of LLM inference for robot control.

\clearpage
\bibliographystyle{ACM-Reference-Format}
\bibliography{bib/abr-short,bib/typefly,bib/zhong}

% \clearpage
% \onecolumn
% \input{sections/appendix}
\end{document}